\documentclass[10pt,journal,compsoc]{IEEEtran}

\ifCLASSOPTIONcompsoc
  \usepackage[nocompress]{cite}
\else
  \usepackage{cite}
\fi

\usepackage{amsmath}
\usepackage[ruled,longend]{algorithm2e}

\usepackage{graphicx}
\usepackage{amsmath,amssymb} 
\usepackage{color}

\usepackage{multirow}

\usepackage{amsmath}

\usepackage{wrapfig}
\usepackage{float}
\usepackage{breqn}

\usepackage{subfigure}
\usepackage{booktabs}

\usepackage[misc]{ifsym}

\begin{document}
%
\title{Dual Attention-in-Attention Model for Joint Rain Streak and Raindrop Removal}

\author{Kaihao Zhang,
        Dongxu Li,
        Wenhan Luo,
        and Wenqi Ren
        \IEEEcompsocitemizethanks{ \IEEEcompsocthanksitem Kaihao Zhang and Dongxu Li are with the College of Engineering and Computer Science, Australian National University, Canberra, ACT, Australia. E-mail: \{kaihao.zhang@anu.edu.au; dongxu.li@anu.edu.au\} \protect\\

\IEEEcompsocthanksitem W. Luo is with the Tencent AI Laboratory, Shenzhen 518057, China.
E-mail: whluo.china@gmail.com \protect \\

\IEEEcompsocthanksitem W. Ren is with State Key Laboratory of Information Security, Institute of Information Engineering, Chinese Academy of Sciences, Beijing, 100093, China. E-mail: rwq.renwenqi@gmail.com. \protect


}

\thanks{Manuscript received April 19, 2005; revised August 26, 2015.}}

%
%

\markboth{Journal of \LaTeX\ Class Files,~Vol.~14, No.~8, August~2015}%
{Shell \MakeLowercase{\textit{et al.}}: Bare Advanced Demo of IEEEtran.cls for IEEE Computer Society Journals}
%



\IEEEtitleabstractindextext{%
\begin{abstract}

Rain streaks and raindrops are two natural phenomena, which degrade image capture in different ways. Currently, most existing deep deraining networks take them as two distinct problems and individually address one, and thus cannot deal adequately with both simultaneously. To address this, we propose a Dual Attention-in-Attention Model (DAiAM) which includes two DAMs for removing both rain streaks and raindrops. Inside the DAM, there are two attentive maps - each of  which attends to the heavy and light rainy regions, respectively, to guide the deraining process differently for applicable regions. In addition, to further refine the result, a Differential-driven Dual Attention-in-Attention Model (D-DAiAM) is proposed with a ``heavy-to-light" scheme to remove rain via addressing the unsatisfying deraining regions. Extensive experiments on one public raindrop dataset, one public rain streak and our synthesized joint rain streak and raindrop (JRSRD) dataset have demonstrated that the proposed method not only is capable of removing rain streaks and raindrops simultaneously, but also achieves the state-of-the-art performance on both tasks. 

\end{abstract}

}

\maketitle

\IEEEdisplaynontitleabstractindextext

%
\IEEEpeerreviewmaketitle

\section{Introduction}
\label{sect:intorduction}

As one of the commonest weather phenomena, rain causes visibility degradation and destroys the performance of many computer vision systems, \textit{e.g.}, object detection \cite{girshick2015fast,he2017mask}, outdoor surveillance \cite{zheng2015scalable,han2005individual} and autonomous driving \cite{yang2019drivingstereo,li2019gs3d}. Rain removal is to restore clean images from rainy ones, which is an important problem in computer vision field and still challenging due to its various types (\textit{i.e.}, rain streaks and raindrops), and different intensities (\textit{i.e.}, heavy and light rain).

\begin{figure}[!tb]
  \centering
  \subfigure[Input rainy images]{
    \label{idea:a}
    \includegraphics[width=0.48\linewidth ]{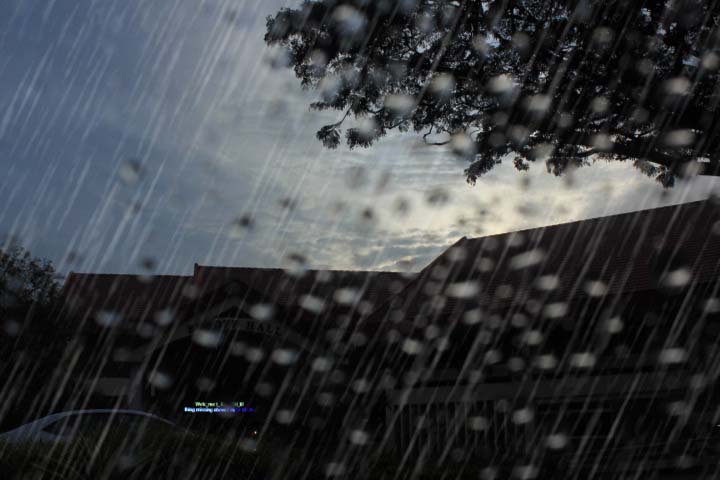}}
  \subfigure[Rainy regions]{
    \label{idea:b}
    \includegraphics[width=0.48\linewidth ]{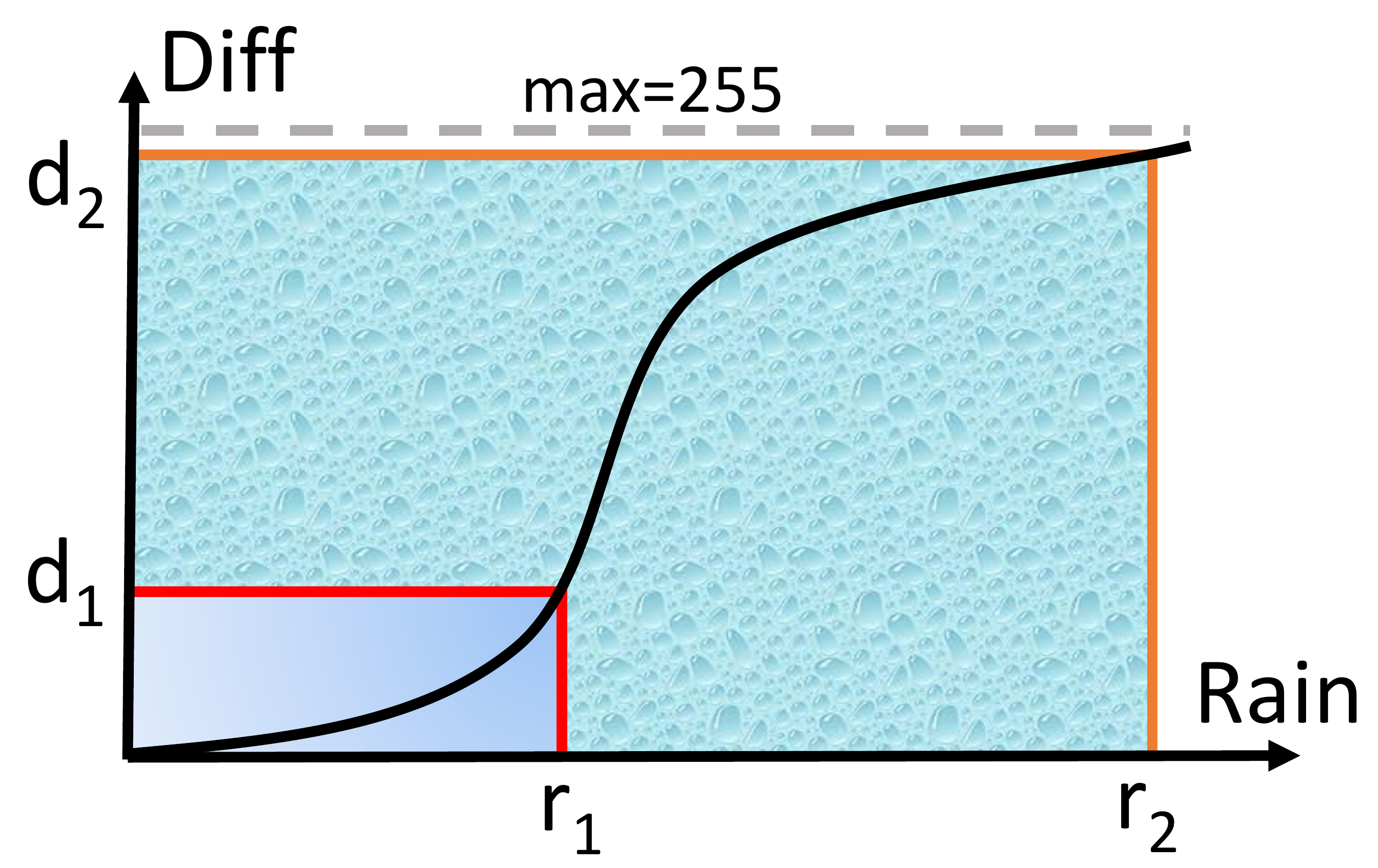}}
    \subfigure[Attention: rain streaks]{
    \label{idea:c}
    \includegraphics[width=0.48\linewidth ]{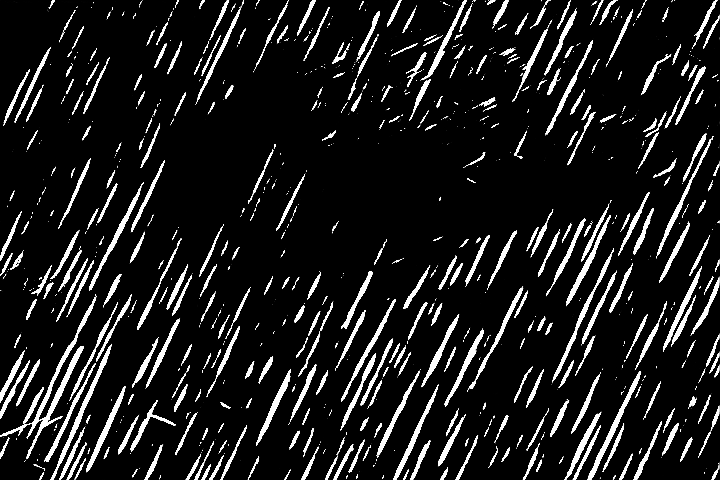}}
  \subfigure[Attention: raindrops]{
    \label{idea:d}
    \includegraphics[width=0.48\linewidth ]{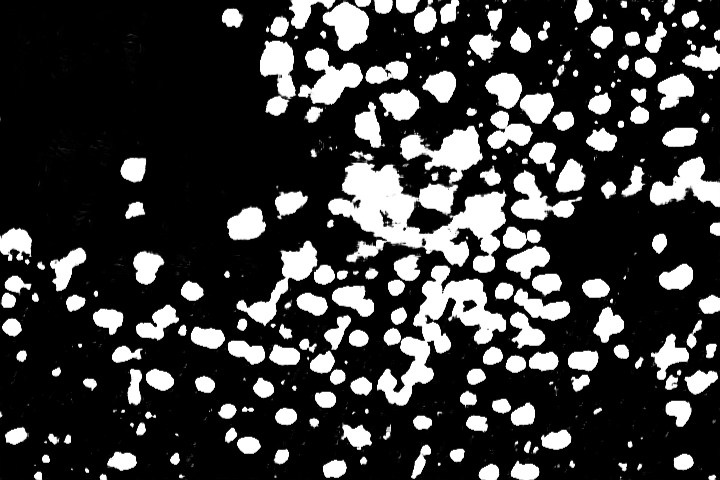}}
    \subfigure[DAM(odd)]{
    \label{idea:e}
    \includegraphics[width=0.48\linewidth ]{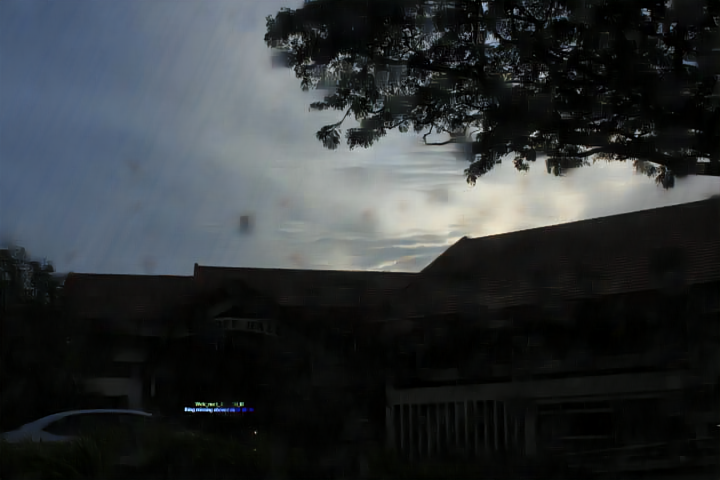}}
  \subfigure[DAM(dual)]{
    \label{idea:f}
    \includegraphics[width=0.48\linewidth ]{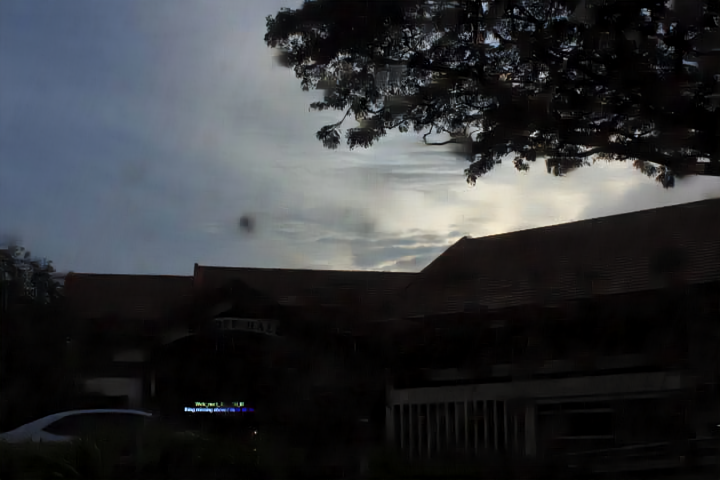}}
\caption{Analyses and deraining results. (a) is an input rainy image. (b) describes the relationship of the rain intensity and the difference between rainy and clean images. (c) and (d) are the generated attention maps for rain streaks and raindrops, respectively. (e) and (f) are the deraining results of the proposed DAM with odd attention and dual attention, respectively.}
  \label{idea}
\end{figure}

In the last decade, a set of methods have been proposed for rain removal. For rain streak removal, some methods model the physical characteristics of rain and generate sharp version with various image priors \cite{sun2014exploiting,kang2011automatic,chen2013generalized,zhang2006rain}. we have also witnessed significant progress of deep learning based methods \cite{fu2017clearing,fu2017removing,li2018recurrent,yang2017deep,zhang2018density}. Some others focus on raindrop removal via detecting and removing raindrop using multiple images or single image \cite{roser2009video,roser2010realistic,roser2010realistic,eigen2013restoring,qian2018attentive}. Despite of the achieved promising performance, there still exist major challenges in rain removal:

\begin{itemize}
\item
Rain streaks and raindrops are two related but different types. The rain streaks lead to the occlusion of objects and scene, while raindrops can cause change of shape. In the real world, both of them often appear simultaneously. However, most deep learning based deraining methods and datasets typically focus on one of them.

\item
As Fig. \ref{idea:b} shows, the pixel difference between clean and rainy images increases as the rain becomes heavier. Previous attention based methods use a fixed threshold $d_1$ to determine whether a pixel is part of rain regions. These methods focus only on the top-right heavy rainy region and ignore the bottom-left light rainy region. In this case, the efficacy of attention mechanism will be restricted if $d_1$ is set inappropriately large or small.

\item
For many cases like heavy rain, the current rain removal methods can remove rain to some extent and generate a derained image with less rain. However, it is difficult to further improve the performance by simply modifying the structure of deep networks like increasing the depth.
\end{itemize}

To address the first and second problems, a new framework which exploits the cues from different types of rain is proposed. Specially, we propose a Dual Attention-in-Attention Model, termed as \textbf{DAiAM}, to remove rain streaks and raindrops, simultaneously. It contains two branches, corresponding to two Dual Attention Model (\textbf{DAM}). Each DAM removes one type of rain via simultaneously focusing on different rain intensities. Different from previous attention-based deraining methods, which learn only the attention map of heavy rain regions (top-right regions in Fig. \ref{idea:b}), an advantage of the DAM is that it also pays attention to the light rain regions (bottom-left regions in Fig. \ref{idea:b}). One pair of heavy-rain-aware and light-rain-aware attention maps is generated to help remove rain from multiple regions.
As such, the proposed method avoids the negative effects from unsuitable thresholds. Fig. \ref{idea:e} and \ref{idea:f} show the attention maps for rain streaks and raindrops, respectively.

For the third challenge, a Differential-driven Dual Attention-in-Attention Model (\textbf{D-DAiAM}), is proposed based on a ``\textit{heavy-to-light}" scheme. The input rainy images and output derained images from DAiAM are processed with the proposed differential-driven module, guiding the learning of the following DAiAM to further remove rain with different intensities or different types. 

In order to evaluate the performance of the proposed method on rain streak and raindrop removal, a joint rain streak and raindrop dataset (\textbf{JRSRD}), is built. The rain streaks and raindrops often happen simultaneously, thus evaluating methods in this scenario is necessary to verify the performance of different methods in the wild.

The contributions of this work can be summarized as: 
1) To address the problem of joint rain streak and raindrop removal, a dual attention-in-attention model (DAiAM), is proposed to remove two variations of rain. 
2) Inside DAiAM, there are two well-designed DAMs, which focus on local regions with different rainy intensities. The generated intensity-aware attention maps enable better removal of rain in multiple regions.
3) D-DAiAM is proposed to alleviate the limitation of increasing depth and width of deraining methods, and thus improve the image quality. 
4) A new JRSRD dataset of both rain streaks and raindrops is built. We compare the proposed method with current deraining methods. Experimental results show that the proposed method achieves not only the state-of-the-art performance on public rain streak dataset and raindrop dataset, but also consistently better results on images with both rain streaks and raindrops.

\section{Related Works}

Our work is an attempt for jointly addressing the rain streak and raindrop removal based on attention mechanism. The following is a brief review of related works on rain streak removal, raindrop removal, as well as attention mechanism, respectively.

\subsection{Rain Streak Removal}
Traditional methods design hand-crafted priors to remove rain streaks \cite{barnum2010analysis,kang2011automatic,huang2013self,luo2015removing,li2016rain,chang2017transformed,zhu2017joint,zhu2020learning,hu2021single,wang2020rethinking}. Kang \textit{et al.} \cite{kang2011automatic} use a bilateral filter to decompose an image into the low- and high-frequency parts, which are then decomposed into different components by performing dictionary learning and sparse coding. Similarly, Huang \textit{et al.} \cite{huang2013self} present a method to first learn an over-complete dictionary from the image high spatial frequency parts and then perform unsupervised clustering on the dictionary atoms. Zhu \textit{et al.} \cite{zhu2017joint} use a joint optimization process with three image priors to remove rain-streak details.

Recently, deep learning achieves significant success in low-level vision tasks \cite{johnson2016perceptual,zhang2018adversarial,zhang2020deblurring,zheng2021t,zhang2020every,zhang2021deep}, which also include rain streak removal \cite{fu2017clearing,fu2017removing,yang2017deep,zhang2018density,li2018recurrent,zhang2019image,zhang2021beyond,zhang2020beyond}. Fu \textit{et al.} \cite{fu2017removing} propose a deep network to remove background interference and focus on the structure of rain based on prior knowledge. Zhang \textit{et al.} \cite{zhang2018density} introduce a DID-MDN model to jointly estimate rain density and remove rain. Li \textit{et al.} \cite{li2018recurrent} propose a deep convolutional and recurrent neural network for deraining. To make the derained images more realistic, Zhang \textit{et al.} \cite{zhang2019image} introduce a CGAN-based model with additional regularization. Wang \cite{wang2020model} explore the intrinsic prior structure of rain streaks and then propose a novel interpretable network to remove the rain streaks from rainy images. 
Li \textit{et al.} \cite{li2021comprehensive} propose a comprehensive benchmark analysis of several single image deraining networks. Zhu \textit{et al.} \cite{zhu2020learning} and Hu \textit{et al.} \cite{hu2021single} introduce two non-local networks to improve the performance of image deraining. Wang \textit{et al.} \cite{wang2020rethinking} rethink about the image deraining and reformulate rain streaks as transmission medium together with vapors to address the problem of image deraining.

In addition, there still exists some video-based rain steak removal methods \cite{li2018video,liu2018d3r,liu2018erase,chen2018robust,yang2019frame,zhang2021enhanced}. Specially, Chen \textit{et al.} \cite{chen2018robust} use a super-pixel segmentation scheme to help restore clean frames via a robust deep CNN. Liu \textit{et al.} \cite{liu2018erase} remove rain streak via classifying all pixels. More recently, Yang \textit{et al.} \cite{yang2019frame} introduce a two-stage recurrent network to capture the motion consistency to remove rain streaks.

\begin{figure*}[!tb]
  \centering
\includegraphics[width=0.9\linewidth ]{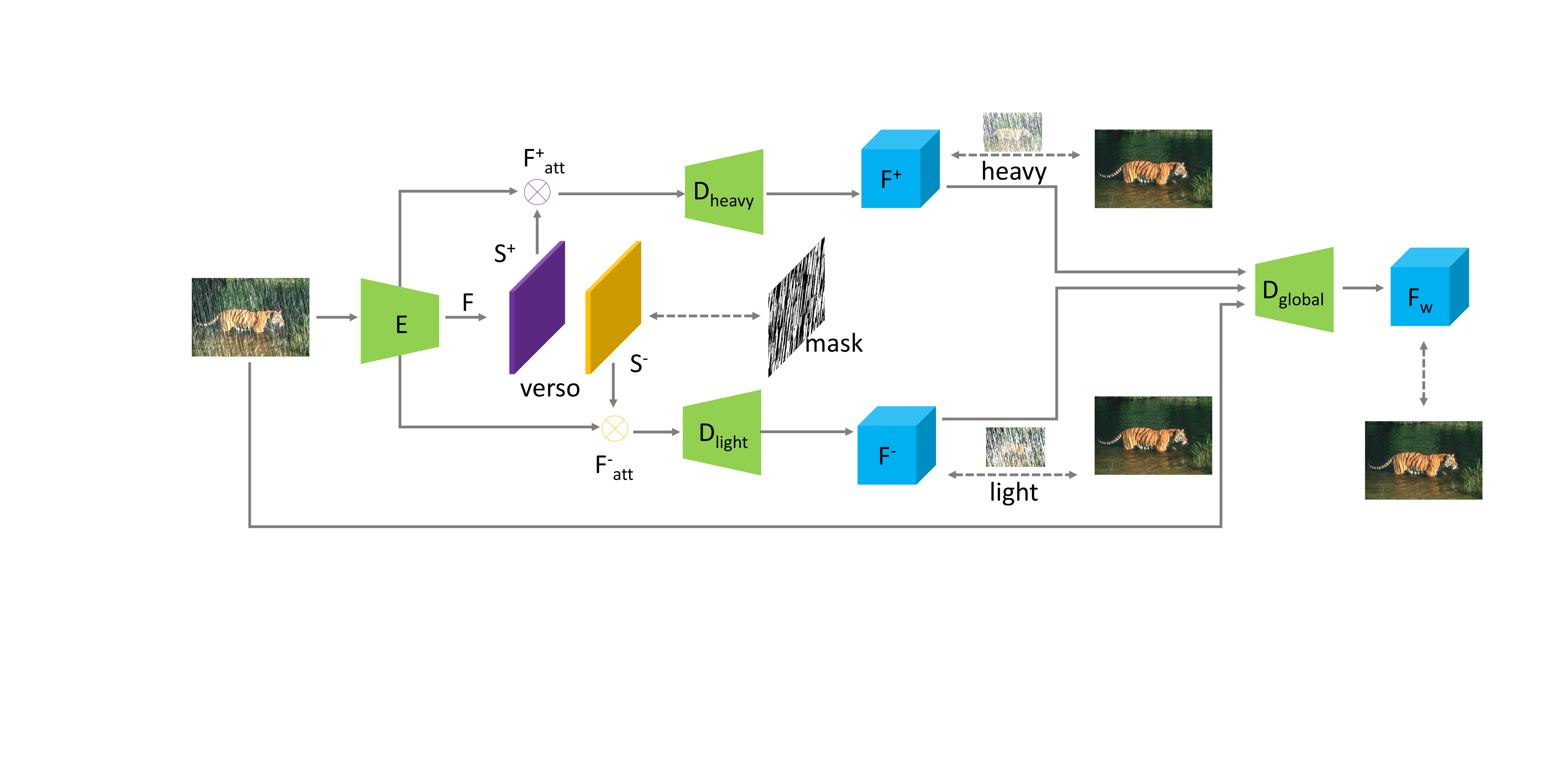}
\caption{The framework of DAM for image deraining. It contains three main branches, \textit{i.e.}, heavy-rain branch, light-rain branch and full-image branch. The dual attention sub-network in the middle is utilized to generate a pair of heavy-rain-aware and light-rain-aware maps to pointedly remove rain from different regions. The original rainy image and the intermediate results are then concatenated to generate the final deraining image.}
\label{framework}
\end{figure*}

\subsection{Raindrop Removal}
Most methods for rain streak removal are not directly applicable for raindrop removal. Therefore, many methods are proposed like raindrop detection and removal \cite{kurihata2005rainy,roser2009video,roser2010realistic,yamashita2005removal,yamashita2009noises,you2015adherent,eigen2013restoring,quan2019deep,alletto2019adherent,hao2019learning}. 
Specially, Kurihata \textit{et al.} \cite{kurihata2005rainy} use PCA to learn the shape of raindrops, which are then utilized to match rainy regions. Yamashita \textit{et al.} \cite{yamashita2005removal} introduce a method based on the stereo measurement and disparities between stereo image pair. Position of raindrops can be detected. Finally, sharp image can be obtained by replacing raindrop regions. Roser \textit{et al.} \cite{roser2009video} propose a method to perform monocular raindrop detection. \cite{you2015adherent} introduces a method to exploit local spatio-temporal cue for video raindrop removal. They first model and detect adherent raindrops, then remove them and restore the images. More recently, there are many methods using CNN for single image raindrop removal \cite{eigen2013restoring,qian2018attentive}, which are trained with pairs of raindrops and corresponding sharp images. Quan \cite{quan2019deep} propose a CNN-based method to restore an image taken through glass window in rainy weather via using shape-driven attention and channel re-calibration.

Almost all existing methods dissever the two tasks and focus on either rain streaks or raindrops \cite{li2019single}. Meanwhile, most datasets typically contain only one kind of rain. Given that the two phenomenons usually appear simultaneously in the real world, a new dataset including raindrops and rain streaks is built in this paper. 

\subsection{Attention Mechanism}
The visual attention model is effective in understanding image. It has achieved great success in tasks like object recognition \cite{ba2014multiple,gregor2015draw,xiao2015application}, image captioning \cite{xu2015show,you2016image} and saliency detection \cite{chen2018reverse,lv2021simultaneously,li2021uncertainty,mao2021transformer}. For example, Ba \textit{et al.} \cite{ba2014multiple} use an attention mechanism to help their model decide where to focus its computation and thus propose a new method to train their object recognition model. Xu \textit{et al.} \cite{xu2015show} propose an attention based approach which can automatically learn to describe the content of images. Chen \textit{et al.} \cite{chen2018reverse} introduce a reverse attention module for salient object detection via guiding residual learning in a top-down manner. For deep image deraining, there are also some methods of attention mechanism \cite{yang2017deep,wang2019spatial}. They utilize a threshold value to classify the regions of input rainy images into two classes like ``rain" or ``no-rain", and then derive a spatial attention map to remove rain. As discussed, unsuitable thresholds cause errors, and restrict the potential of attention mechanism.

\section{Method}

We first take rain streak removal as an example to introduce the architecture and learning details of DAM. Then we represent DAiAM (Sec. \ref{sec_DAIAM}) to jointly remove rain streaks and raindrops. Finally, a D-DAiAM framework (Sec. \ref{BDAM}) is discussed to overcome the limitation of single model.

\subsection{Overall Architecture of DAM} 
\label{architecture}
The overall architecture of the proposed \textbf{DAM} is shown in Fig. \ref{framework}. A rainy image is fed into DAM to learn two attention maps, \textit{i.e.}, heavy-rain-aware and light-rain-aware maps. The heavy-rain-aware map learns the attention which indicates the regions with heavy rain, and the light-rain-aware map represents the regions with light rain (Sec. \ref{dual_attention_map}). 

Different from other deraining methods which directly concatenate the attention maps to generate final images, we produce two different kinds of intermediate results by two sub-networks in Sec. \ref{attention_derianing}. The two attention maps provide not only attention to generate the final global deraining image, but also the reference to evaluate the performance of two sub-networks of DAM. Finally, the intermediate results concatenated with the input rainy image are put into a global decoder to generate the deraining image.

\subsection{Dual Intensity-Aware Maps}
\label{dual_attention_map}
In general, the DAM takes input images and produce weighting maps to focus on different spatial regions of images. By doing so, different sub-networks can exactly focus on different spatial regions that contribute most for differentiated image deraining. 
Specially, the proposed DAM take rainy images as input to capture the features $F$ from the first-step encoder $E$. Then the feature maps are fed into two attention sub-networks to generate heavy-rain-aware and light-rain-aware maps, respectively. The heavy-rain-aware map $S^{+}$ can be defined as:
\begin{equation}
\label{heavy}
S^{+} = g(W*F+b) \ ,
\end{equation}
where $*$, $W$ and $b$ denote respectively convolution, convolution filters and biases. $g$ is the sigmoid function.

\begin{figure*}[!tb]
  \centering
\includegraphics[width=0.8\linewidth ]{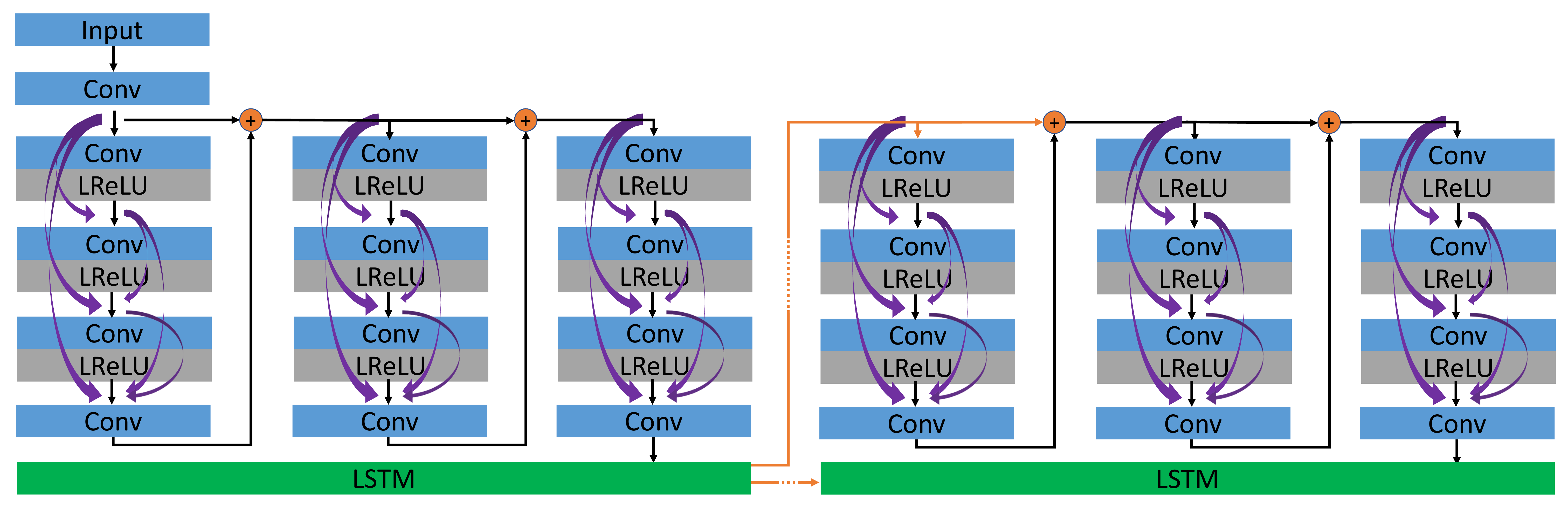}
\caption{The architecture of $E$. The detail of the architecture is shown in Table. \ref{table_G}.The arrows in the ResBlock mean the residual learning.}
\label{figure_E}
\end{figure*}

Fig. \ref{figure_E} shows the architecture of $E$, and Table \ref{table_G} presents the detailed configurations. As Fig. \ref{figure_E} shows, $E$ is a structure of a recurrent network 
consisting of one CNN layer, three residual block and one LSTM layer. 
The output is the feature $F$ captured from $E$. The attention map can be generated based on Eq. (1) in main submission.

\begin{figure*}[!tb]
  \centering
\includegraphics[width=0.8\linewidth ]{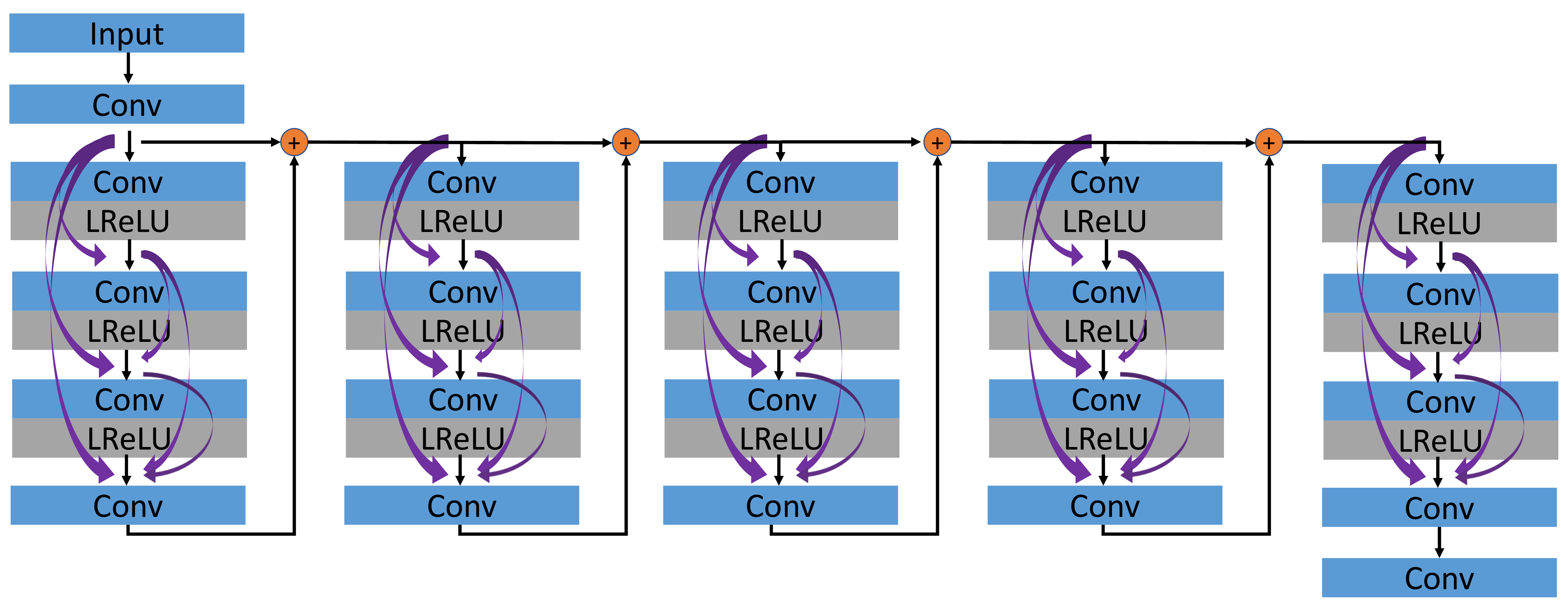}
\caption{The architecture of $D_{heavy}$ and $D_{light}$. The detail of the architecture is shown in Table. \ref{table_D}. The arrows in the ResBlock mean the residual learning.}
\label{figure_D}
\end{figure*}

\begin{table}[!tb]
  \centering
  \caption{Configurations of the encoder $E$.}
    \begin{tabular}{cccccc}
    \toprule
    layers & kernel size & output & operations & skip connection\\
    \midrule
     CNN     & $3 \times 3 $  & 32 & - & ResBlock1 \\
     \midrule
     ResBlock1     & $3 \times 3 $  & 32 & LReLU & ResBlock2 \\
     ResBlock2     & $3 \times 3 $  & 32 & LReLU & ResBlock3 \\     
     ResBlock3     & $3 \times 3 $  & 32 & LReLU & - \\
     \midrule
     LSTM    & $3 \times 3 $  & 32 & Tanh & - \\
    \bottomrule
    \end{tabular}
  \label{table_G}
\end{table}

Then we can similarly generate the light-rain-aware map based on Eq. \eqref{heavy}. The heavy and light rain regions are a pair of complementary regions. Thus a constraint of them is set as: 
\begin{equation}
S^{+} + S^{-} = 1\,.
\end{equation}

The two attention maps are two weighting maps which denote different region-aware attentions from the input features. Based on them, it is easy for the following sub-networks to pay attention to different regions and obtain different outputs. The operation to obtain the different features based on the two attention maps can be represented as,
\begin{equation}
\label{f_heavy}
F^{+}_{att} = F \otimes S^{+}\, ,
\end{equation}
\begin{equation}
\label{f_light}
F^{-}_{att} = F \otimes S^{-}\, ,
\end{equation}
where $\otimes$ denotes the channel-wise Hadamard matrix operation. $F^{+}_{att}$ and $F^{-}_{att}$ have the same size as $F$ but are two re-weighted features by the two attention maps to focus on heavy-rain and light-rain regions, respectively. The $S^{-}$ is the light-rain-aware attention map, where light-rain regions have higher weights and the heavy-rain regions have lower values. In order to guarantee that $S^{+}$ learns the heavy-rain regions, we develop another constraint to make it focus on the heavy-rain regions and thus simultaneously push $S^{-}$ to learn the light-rain regions. The loss function with this constraint is represented as
\begin{equation}
\mathcal{L}_{att} = \sum_{x=1}^{X} \sum_{y=1}^{Y} M_{(x,y)} - S^{+}_{(x,y)}\, ,
\end{equation}
where $M$ is the rain-aware mask. $X$ and $Y$ are the width and height of the input features. Different from the previous methods \cite{yang2017deep,wang2019spatial}, which use a binary mask to represent the rain and no-rain regions, we apply a ``soft" manner. Specially, we calculate the difference of images between the rainy and non-rainy versions and then normalize to the range between 0 and 1. This not only denotes whether the regions are rainy or not, but also represents the intensity of rain. In this way, we can avoid the negative effects caused by inappropriate thresholds and binary masks. Based on the above mechanism, two different attention maps are obtained with focus on heavy-rain and light-rain regions, respectively.

Namely, during the training stage, if the datasets do not provide rain streak and raindrops maps, we can generate them via calculating the difference between a rainy image and its corresponding clean image. Heavier rain corresponds to greater values in the map. This works well in the practice.

\begin{figure*}[!tb]
  \centering
\includegraphics[width=0.6\linewidth ]{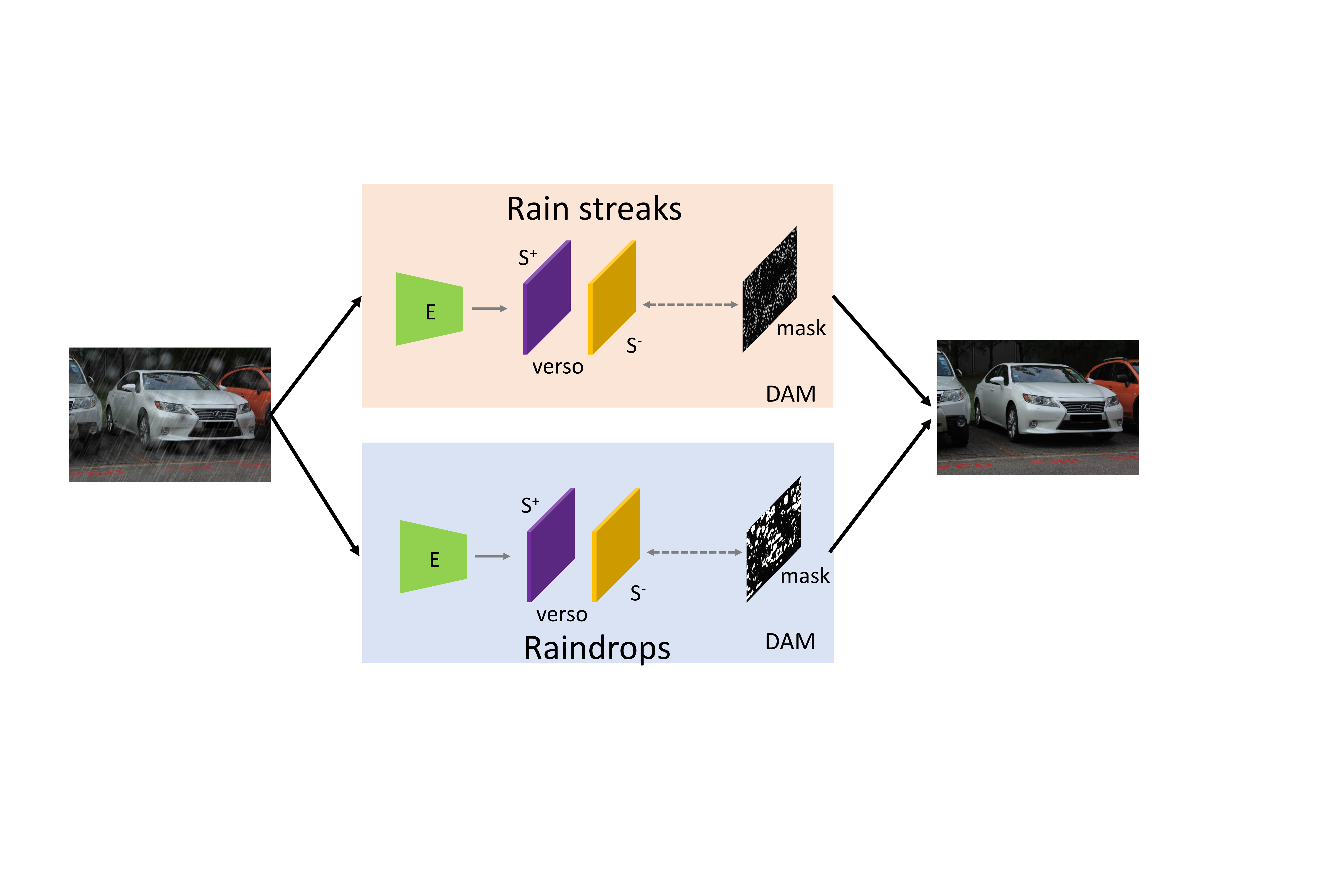}
\caption{The framework of DAiAM for joint rain streak and raindrop removal. DAiAM takes a rainy image as input to capture attention maps for rain streaks and raindrops via two DAMs. Then the outputs of them are concatenated to generate final deraining result.}
\label{framework_DAIAM}
\end{figure*}

\subsection{Attentive Deraining from Regional and Global Levels}
\label{attention_derianing}
After the two attention maps are generated, we can improve the performance of deep deraining networks with them as reference. Specially, the attended features with the heavy-rain-aware attention map $S^{+}$ and light-rain-aware attention map $S^{-}$ are sent into two decoder networks to reconstruct two different deraining images with focus on different regions. The learning process can be defined as:
\begin{equation}
\label{L_heavy}
\mathcal{L}_{heavy} = I_{c} - D_{heavy}(F^{+}_{att}, I_i) \ ,
\end{equation}
\begin{equation}
\label{L_light}
\mathcal{L}_{light} = I_{c} - D_{light}(F^{-}_{att}, I_i) \ ,
\end{equation}
where $I_{c}$ denotes the clean image and $I_{i}$ is the input rainy image. The decoder networks $D_{heavy}$ and $D_{light}$ generate two deraining images, and the attentions of them are different. $\mathcal{L}_{heavy}$ specially constrains the network $D_{heavy}$ to mainly focus on the heavy-rain regions but consider less the light-rain regions due to the weighting values from $S^{+}$. The $\mathcal{L}_{light}$ pushes the $D_{light}$ to remove rain from light regions. Finally, both of the intermediate deraining images are concatenated with the original rainy image to generate the final deraining image via a global decoder, denoted as:
\begin{equation}
I_{o} = D_{global}(F^{+}, F^{-}, I_{i}) \ ,
\end{equation}
where $I_{o}$ is the derained image. We use MSE to update the model as
\begin{equation}
\mathcal{L}_{global} = \sum_{x=1}^{X} \sum_{y=1}^{Y} I_{c(x,y)} - I_{o(x,y)}\, .
\end{equation}

The final loss function of the DAM contains $\mathcal{L}_{att}$, $\mathcal{L}_{heavy}$, $\mathcal{L}_{light}$ and $\mathcal{L}_{global}$, which is defined as,
\begin{equation}
\mathcal{L}_{DAM} = \alpha \cdot \mathcal{L}_{att} + \beta_1 \cdot \mathcal{L}_{heavy} + \beta_2 \cdot \mathcal{L}_{light} + \mathcal{L}_{global} \ ,
\end{equation}
where $\alpha$, $\beta_1$ and $\beta_2$ are three parameters to balance different loss functions, respectively.

$D_{heavy}$ and $D_{light}$ share a similar recurrent structure, including one CNN layer, five residual blocks and another CNN layer, as Fig. \ref{figure_D} shows. Table \ref{table_D} provides the detailed configurations. 
The $D_{global}$ has a similar architecture, \emph{i.e.}, a recurrent structure of one CNN layer, two residual blocks, and one additional CNN layer. Its network configurations can refer to Table \ref{table_DG}.

\begin{table}[!tb]
  \centering
  \caption{Configurations of the proposed $D_{heavy}$ and $D_{light}$.}
    \begin{tabular}{cccccc}
    \toprule
    layers & Kernel size & output & operations & skip connection\\
    \midrule
     CNN1     & $3 \times 3 $  & 64 & - & ResBlock1 \\
     \midrule
     ResBlock1     & $3 \times 3 $  & 64 & LReLU & ResBlock2 \\
     ResBlock2     & $3 \times 3 $  & 64 & LReLU & ResBlock3 \\     
     ResBlock3     & $3 \times 3 $  & 64 & LReLU & ResBlock4 \\
     ResBlock4     & $3 \times 3 $  & 64 & LReLU & ResBlock5 \\
     ResBlock5     & $3 \times 3 $  & 64 & LReLU & - \\
     \midrule
      CNN2     & $3 \times 3 $  & 3 & - & - \\
    \bottomrule
    \end{tabular}
  \label{table_D}
\end{table}

\begin{table}[!tb]
  \centering
  \caption{Configurations of the proposed $D_{global}$.}
    \begin{tabular}{cccccc}
    \toprule
    layers & Kernel size & output & operations & skip connection\\
    \midrule
     CNN1     & $3 \times 3 $  & 64 & - & ResBlock1 \\
     \midrule
     ResBlock1     & $3 \times 3 $  & 64 & LReLU & ResBlock2 \\
     ResBlock2     & $3 \times 3 $  & 64 & LReLU & - \\
     \midrule
      CNN2     & $3 \times 3 $  & 3 & - & - \\
    \bottomrule
    \end{tabular}
  \label{table_DG}
\end{table}

\subsection{Dual Attention-in-Attention Model}
\label{sec_DAIAM}
As discussed above, raindrops and rain streaks are two different rain types and usually appear simultaneously in the real world. In this case, rain removal becomes a more challenging problem. Previous methods \cite{li2019single} often focus on removing one type of rain from rainy images. To simultaneously remove both of them, a Dual Attention-in-Attention Model, \textbf{DAiAM}, is proposed.
The first attention-in-attention model is responsible for heavy rain raindrops and rain streaks, while the second attention-in-attention model is responsible for light rain raindrops and rain streaks.
As a comparison, in the Dual Attention Model (DAM), the first attention model is responsible for heavy rain raindrops or rain streaks, while the second attention model is responsible for light rain raindrops or rain streaks. Among the Attention-in-Attention model, the first attention is responsible for rain types, while second attention is responsible for the rain intensities.


Fig. \ref{framework_DAIAM} shows the core idea of DAiAM. Image of raindrops and rain streaks is fed into our proposed DAiAM, which has two branches to pay attention to removal of raindrops and rain streaks, respectively. The branch for raindrop removal is similar to the method of removing rain streaks, which is represented in the above based on DAM. The main difference is that the attention loss function $\mathcal{L}_{att}$ is calculated based on the mask of raindrops, rather than rain streaks. In this way, the DAiAM first pays attention to two kinds of rain variations, and then focuses on two kinds of rain intensity in different branches. The final loss function of DAiAM is defined as,
\begin{equation}
\mathcal{L}_{DAiAM} = \mathcal{L}_{streak} + \mathcal{L}_{drop} + \mathcal{L}_{global} \ ,
\end{equation}
where $\mathcal{L}_{drop}$ and $\mathcal{L}_{streak}$ are two loss functions to remove raindrops and streaks, respectively. The loss functions of them are
\begin{equation}
\mathcal{L}_{streak} = \alpha \cdot \mathcal{L}^{streak}_{att} + (\beta_1 \cdot \mathcal{L}^{streak}_{heavy} + \beta_2 \cdot \mathcal{L}^{streak}_{light}) ,
\end{equation}
\begin{equation}
\mathcal{L}_{drop} = \alpha \cdot \mathcal{L}^{drop}_{att} + (\beta_1 \cdot \mathcal{L}^{drop}_{heavy} + \beta_2 \cdot \mathcal{L}^{drop}_{light}) ,
\end{equation}
where $\alpha$, $\beta_1$ and $\beta_2$ are parameters to balance different loss terms. The attention loss function $L^{drop}_{att}$ and $L^{streak}_{att}$ are calculated based on the masks of raindrops and rain streaks, respectively.

Finally, the proposed structure implements the fusion operation of two branches, as is achieved via $D_{global}$. Similar to the method for rain streak removal in Fig. 2, we use a parallel architecture to detect raindrop and then extract features (F{+} and F{-}). All of them are concatenated and then fed into $D_{global}$ to generate final derained images.

\begin{figure}[!tb]
  \centering
\includegraphics[width=0.9\linewidth ]{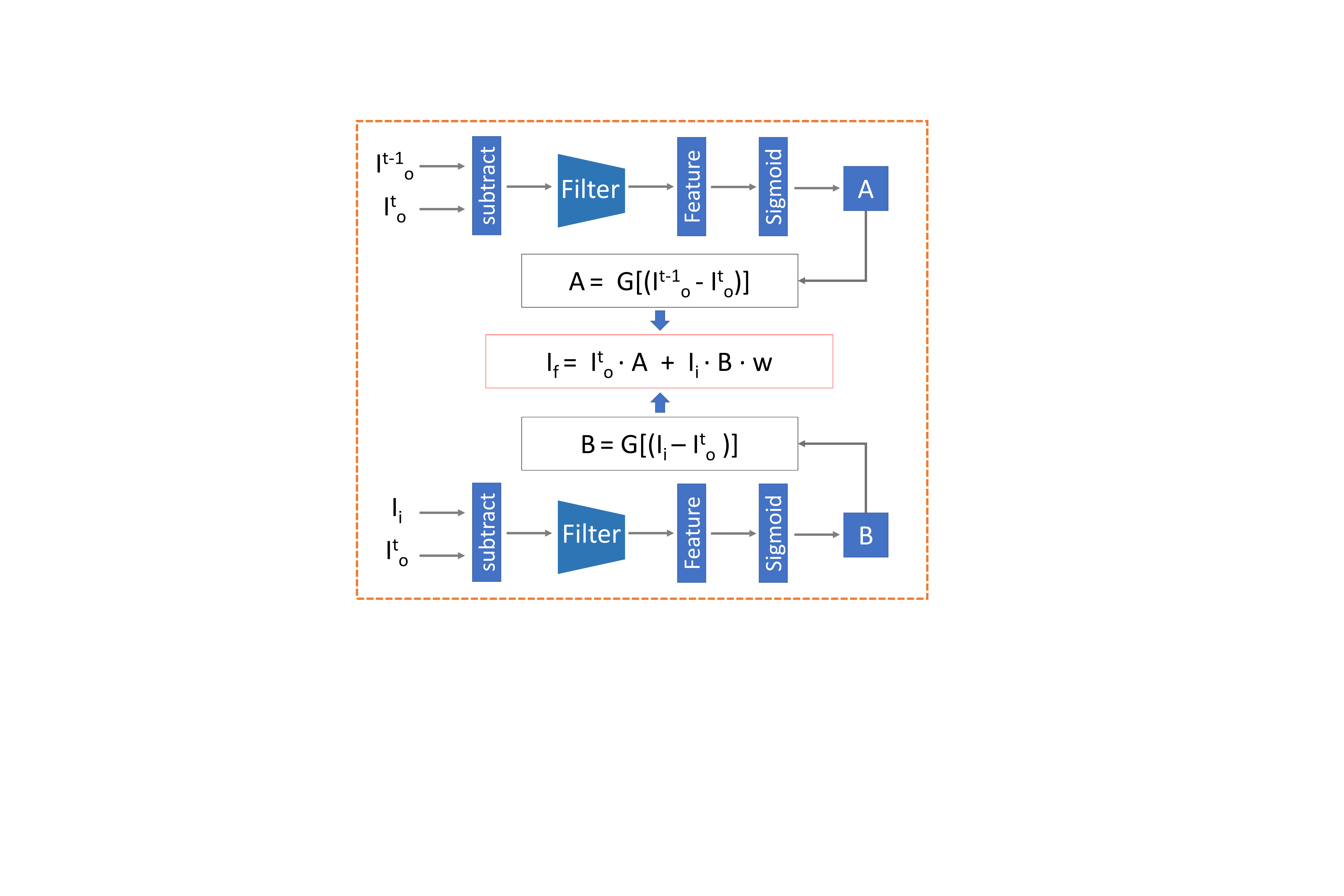}
\caption{The illustration of the differential-driven module. It consists of three streams, \textit{i.e.}, two differential streams and a fusion stream. The FilterNet inside it pointedly selects key regions to help remove rain in the next stage.}
\label{button}
\end{figure}

\subsection{Differential-Driven DAiAM (D-DAiAM)} 
\label{BDAM}

Rain has different intensities and various types.
Images exhibiting both rain streaks and raindrops also pose increasing difficulty of deraining. Deep deraining methods can remove rain to some extent and transfer the heavy-rain images to light-rain ones \cite{hu2019depth,zhang2018density}. However, the performance of a single model is often limited. Simply increasing neural network depth is easy to exhaust the potential and difficult to further improve the performance of rain removal, even for some special heavy rain removal methods \cite{li2019heavy}.

Li \textit{et al.} \cite{li2019single} show that light rainy images are easier to derain. Therefore, we propose a differential-driven dual attention-in-attention model, \textbf{D-DAiAM}, to remove various kinds of rain. Different from most methods \cite{li2019single} which aim to directly derive final deraining images via increasing the depth or width of a single model, we aim to remove heavy rains via transferring heavy rain to light rain and then to no rain in multiple stages. In each stage, we use a DAiAM to generate better visible deraining images and attention information driven by the \textit{differential between the current output and original input}, and the \textit{differential between the current and previous outputs}.

Specifically, this process is conducted via a differential-driven module. As shown in Fig. \ref{button}, we calculate two types of differential. One is the difference between the current output $I^{t}_o$ and the original input $I_i$.
By comparing these two items, the differential is able to guide the following stage to focus on the remaining rainy regions in $I^{t}_o$. The other is the difference between the current and the previous outputs ($I^{t}_o$ and $I^{t-1}_o$). This differential leads the next stage to pay special attention to regions of the current output $I^{t}_o$ which are not handled well in the current stage.

\begin{figure}[tb]
  \centering
  \subfigure[Input rainy image]{
    \label{rainstreaks:a}
    \includegraphics[width=0.48\linewidth ]{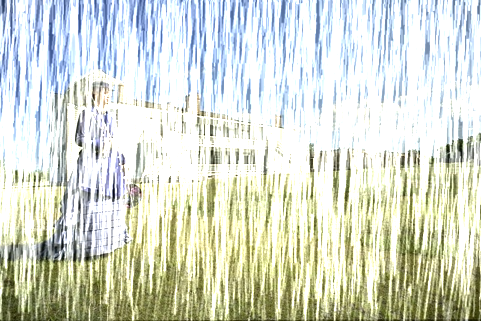}}
  \subfigure[Ground truth]{
    \label{rainstreaks:b}
    \includegraphics[width=0.48\linewidth ]{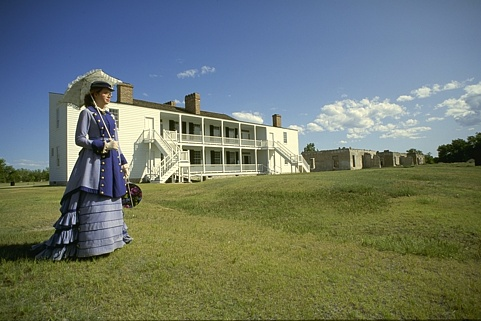}}
    \subfigure[Attention: Ground truth]{
    \label{rainstreaks:c}
    \includegraphics[width=0.48\linewidth ]{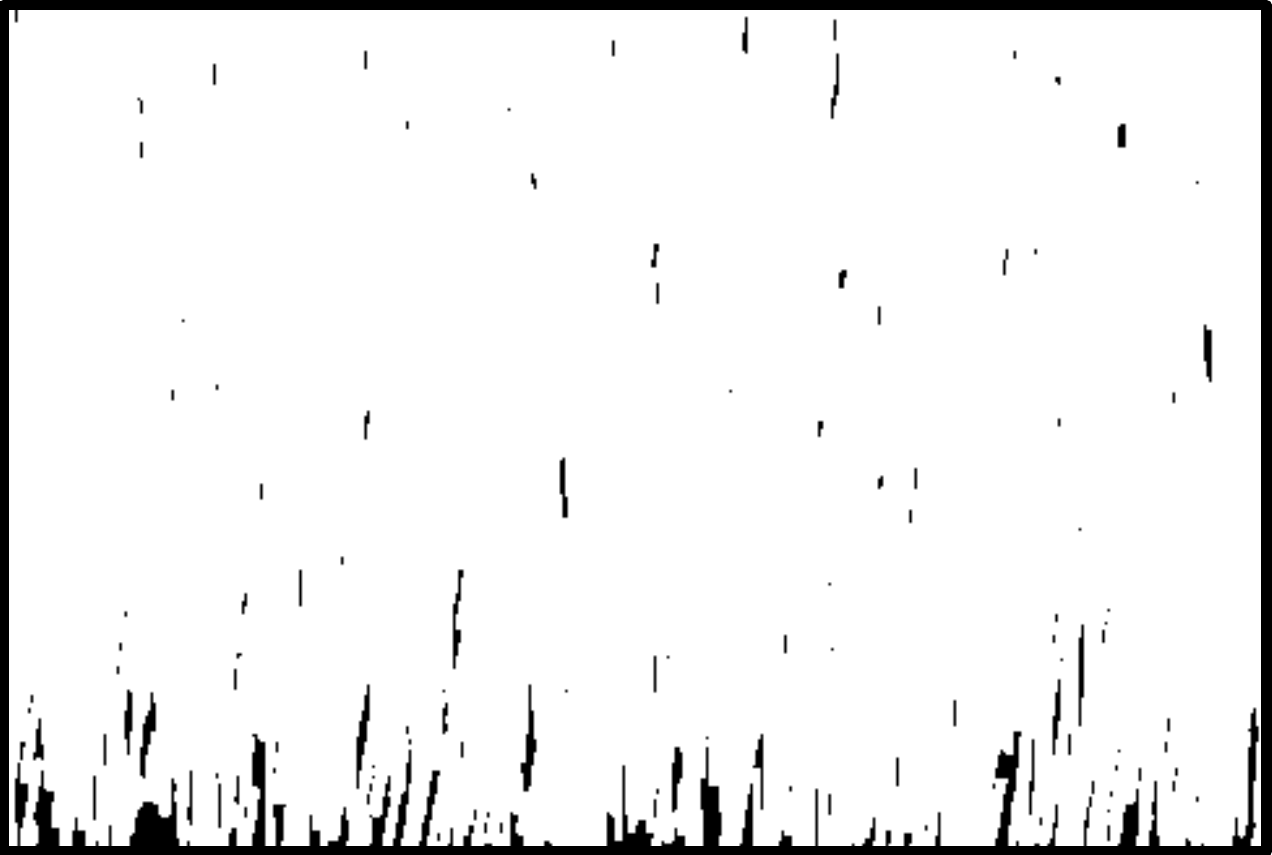}}
  \subfigure[Attention: Ours]{
    \label{rainstreaks:d}
    \includegraphics[width=0.48\linewidth ]{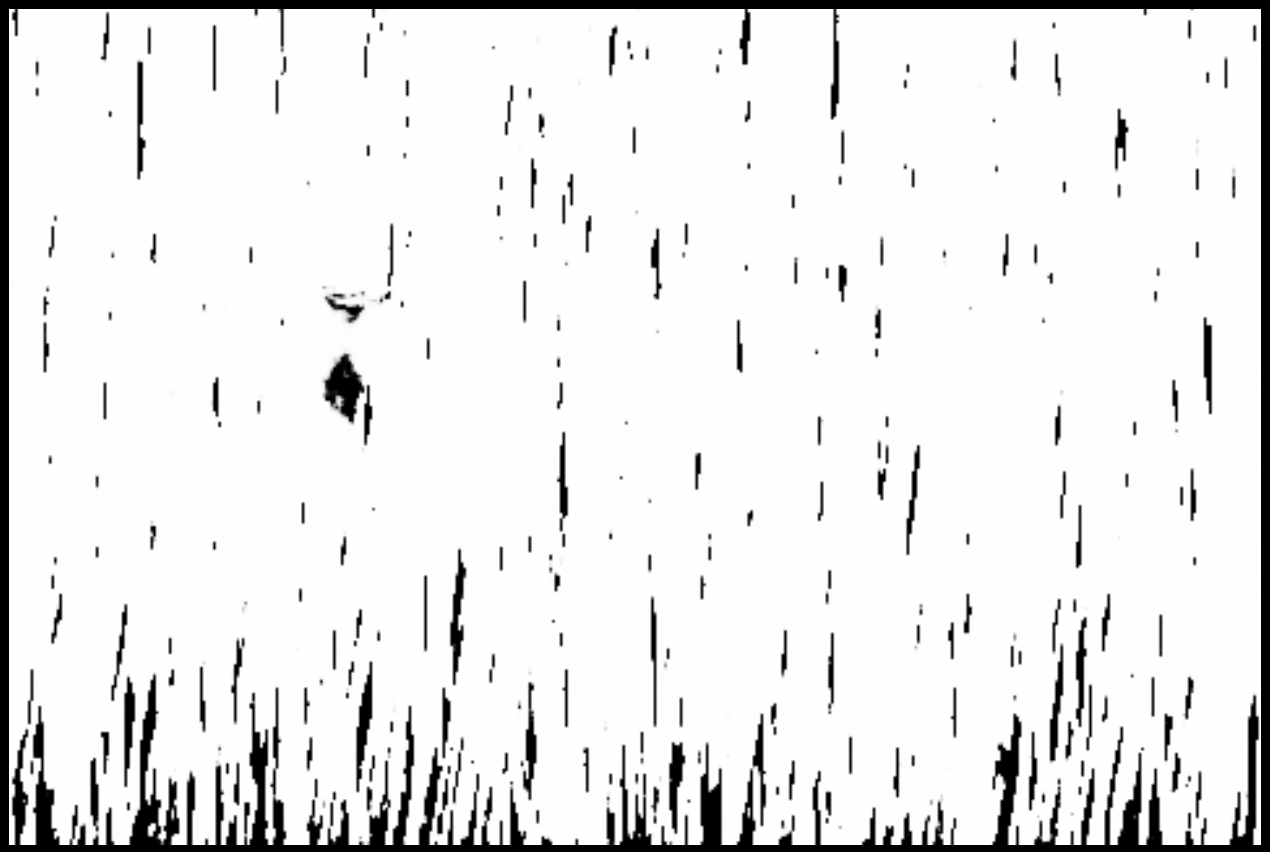}}
     \subfigure[Deraining: PReNet \cite{ren2019progressive}]{
    \label{rainstreaks:e}
    \includegraphics[width=0.48\linewidth ]{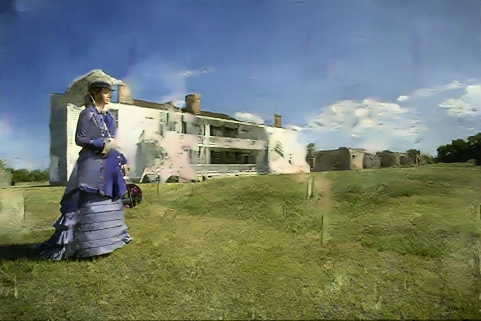}}
  \subfigure[Deraining: Ours]{
    \label{rainstreaks:f}
    \includegraphics[width=0.48\linewidth ]{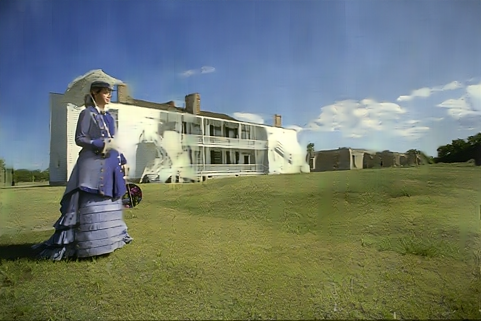}}
\caption{Heavy rain streak removal results of sample images from Rain Streak dataset \cite{yang2017deep}.} 
  \label{figure_rainstreaks}
\end{figure}

\begin{table}[tb]
  \centering 
    \caption{Performance of different model structures on the Rain Streak dataset \cite{yang2017deep} in terms of PSNR and SSIM.}
    \setlength\tabcolsep{5.0pt}
    \begin{tabular}{l |  c c }
    \toprule
    Methods &  PSNR & SSIM \\
    \hline
    GMM \cite{li2016rain} &15.05 &0.425 \\
    DDN \cite{fu2017removing} & 21.92 & 0.764  \\
    RGN \cite{fu2017clearing} & 25.25 & 0.841  \\
    JORDER \cite{yang2017deep} & 26.54 & 0.835  \\
    RESCAN \cite{li2018recurrent} & 28.88 & 0.866  \\
    PReNet \cite{ren2019progressive} & 29.46 & 0.899  \\
    \hline
    \textbf{DAM} & \textbf{29.99} & \textbf{0.905}  \\
    \textbf{D-DAM} & \textbf{30.35} & \textbf{0.907}  \\
    \bottomrule
    \end{tabular}%
    \label{table_raindrops_dataset1}
\end{table}%

Based on these two kinds differential, we employ two \textit{FilterNets} to generate soft maps $A$ and $B$ for our purpose, \textit{i.e.}, the mark of regions needing special attention in the next stage. The FilterNet includes three convolutional layers with $2 \times 2$ kernels to perceive local regions, rather than directly using the input differences. We apply these two soft maps to the original input $I_i$ and the current output $I^{t}_o$ and fuse them, as defined in 
\begin{equation}
I_f = I^{t}_o \otimes A + I_i \otimes B \cdot w,
\end{equation}
where $w$ balances different types of differential.

The coarsest-level DAiAM locates in the begin of D-DAiAM. A latent deraining image is generated at the end of this stage. Even there still exists rain, the generated deraining image exhibits lighter rain. Then, the information from the coarsest level output is addressed by the differential-driven module, and then fed into finer-level network (which has a similar architecture as DAiAM) with deraining images. The final derained image is the output of the last DAiAM.
The objective function to update the D-DAiAM is denoted as:
\begin{equation}
\mathcal{L} = \sum_{t=1}^{N}||I^{t}_{o} - I_c|| \ ,
\end{equation}
where $I^{t}_{o}$ is the derained image in the $t$-th stage and $I_c$ is the ground-truth image.

\section{Experiments}

We first introduce the implementation details. Then the performance of rain streak removal and raindrop removal is compared with the state-of-the-art methods on two public datasets. We develop a new dataset of joint rain streaks and raindrops and test different deraining methods on it. Further, ablation study is carried out to verify the components of our proposal.
Finally, the application of deraining in real-world scenarios is demonstrated.

\begin{figure}[tb]
  \centering
  \subfigure[Input rainy image]{
    \label{raindrops:a}
    \includegraphics[width=0.48\linewidth ]{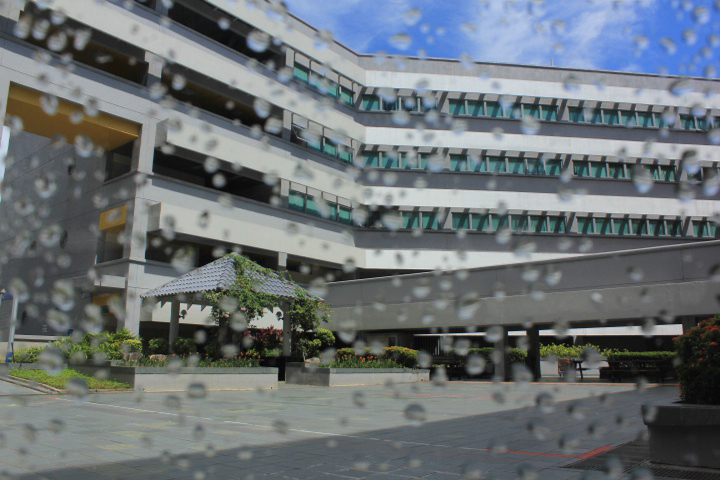}}
  \subfigure[Attention: Ours]{
    \label{raindrops:b}
    \includegraphics[width=0.48\linewidth ]{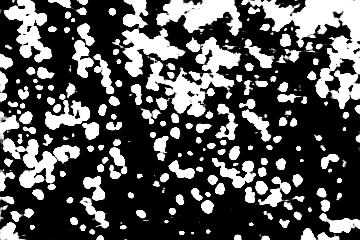}}
     \subfigure[Qian \textit{et al.} \cite{qian2018attentive}]{
    \label{raindrops:c}
    \includegraphics[width=0.48\linewidth ]{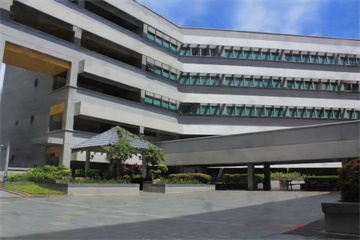}}
  \subfigure[Deraining: Ours]{
    \label{raindrops:d}
    \includegraphics[width=0.48\linewidth ]{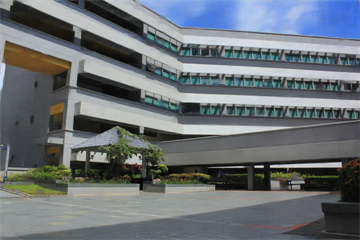}}
\caption{Raindrop removal results on sample images from the Qian \textit{et al.} \cite{qian2018attentive} Raindrop dataset.}
  \label{figure_raindrops}
\end{figure}

\begin{table}[tb]
  \centering 
    \caption{Performance of different model structures on the Raindrop dataset \cite{qian2018attentive} in terms of PSNR and SSIM.}
    \setlength\tabcolsep{5.0pt}
    \begin{tabular}{l |  c c }
    \toprule
    Methods &  PSNR & SSIM  \\
    \hline
    DID-MDN \cite{zhang2018density} & 24.76 & 0.7930 \\
    DDN \cite{fu2017removing} & 25.23 & 0.8366  \\
    JORDER \cite{yang2017deep} & 27.52 & 0.8239  \\
    Qian \textit{et al.} \cite{qian2018attentive} & 30.55 & 0.9023  \\
    Quan \textit{et al.} \cite{quan2019deep} & 30.86 & 0.9263  \\
    Hao \textit{et al.} \cite{hao2019learning} & 30.17 & 0.9128  \\
    \hline
    \textbf{DAM} & \textbf{30.26} & \textbf{0.9137}  \\
    \textbf{D-DAM} & \textbf{30.63} & \textbf{0.9268}  \\
    \bottomrule
    \end{tabular}%
    \label{table_raindrops}
\end{table}%

\subsection{Implementation Details} 

The weights of the proposed networks are initialized with Gaussian distribution with zero mean and a standard deviation of $0.01$. The parameters are updated after a mini-batch of size $4$ in each iteration. In the training stage, $112 \times 112$ patches at random locations of an image are cropped to increase the number of training samples. We also randomly flip training images (horizontally) to further augment the training set. The models are trained under a learning rate which starts with a value of $10^{-4}$ and reduces to $10^{-6}$ after the training has converged. The hyper-parameters $\alpha$, $\beta_1$, $\beta_2$ and $w$ are set as $0.8$, $1.0$, $0.3$ and $0.5$, respectively. To reduce training time, we apply one differential-driven module in our practice. The encoder $E$ contains three residual blocks \cite{he2016deep} and one LSTM layer. $D_{heavy}$ and $D_{light}$ contain one CNN layer, five residual blocks and another CNN layer. $D_{global}$ contains two residual blocks and one CNN layer. The size of all the kernels in this work is set to $3 \times 3$. ReLU function is adopted after convolution operation except the last CNN layer in each structure. 

\subsection{Results on Rain Streak Dataset} 

Yang \textit{et al.} \cite{yang2017deep} build a dataset of heavy rain streaks, named as Rain100H. In order to synthesize heavy rain, they apply two different methods, including the photo-realistic rendering techniques proposed by \cite{garg2006photorealistic} and directly adding simulated sharp line streaks to clear images. The Rain100H dataset consists of $1,800$ and $100$ pairs of images for training and testing, respectively. \cite{ren2019progressive} removes some training images with the same background contents as testing images. Table \ref{table_raindrops_dataset1} reports the comparison results with the state-of-the-art rain streak removal methods, including GMM \cite{li2016rain}, DDN \cite{fu2017removing}, RGN \cite{fu2017clearing}, JORDER \cite{yang2017deep}, RESCAN \cite{li2018recurrent} and PReNet \cite{ren2019progressive}. 
Note that, as the rainy images contain only rain streaks, our full method D-DAiAM degrades as D-DAM in this scenery. Specially, this is only a dual-attention model which focuses on heavy and light rain streak regions. The quantitative results demonstrate the advance of our proposed method over the existing methods. Fig. \ref{figure_rainstreaks} shows the qualitative deraining results and the associated attention maps. Our result is better than that of PReNet \cite{ren2019progressive}. The latent attention map is also close to the ground truth.

\subsection{Results on Raindrop Dataset} 

Qian \textit{et al.} \cite{qian2018attentive} capture $1,119$ pairs of images with different background scenes and raindrops. They use two glasses to model the raindrops. One is clean to capture GT images. The other is sprayed with water to generate corresponding rainy version. The training set and testing set A include $861$ and $58$ pairs, respectively. In order to verify the performance of the propose method, we compare with state-of-the-art deraining methods. As mentioned before, our method becomes D-DAM in this case. Table \ref{table_raindrops_dataset1} presents the results of DID-MDN \cite{zhang2018density}, DDN \cite{fu2017removing}, JORDER \cite{yang2017deep}, Qian \textit{et al.} \cite{qian2018attentive} and ours, respectively. The deraining results and attention maps are provided in Fig. \ref{figure_raindrops}. Both the quantitative and the qualitative results reveal that our method is more advanced.

\begin{figure}[tb]
  \centering
  \subfigure[Input rainy image]{
    \label{jrdrs:a}
    \includegraphics[width=0.48\linewidth ]{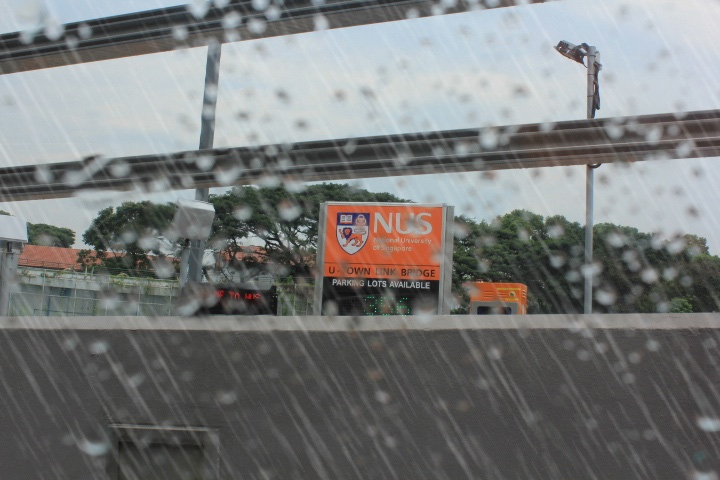}}
  \subfigure[PReNet \cite{ren2019progressive}]{
    \label{jrdrs:b}
    \includegraphics[width=0.48\linewidth ]{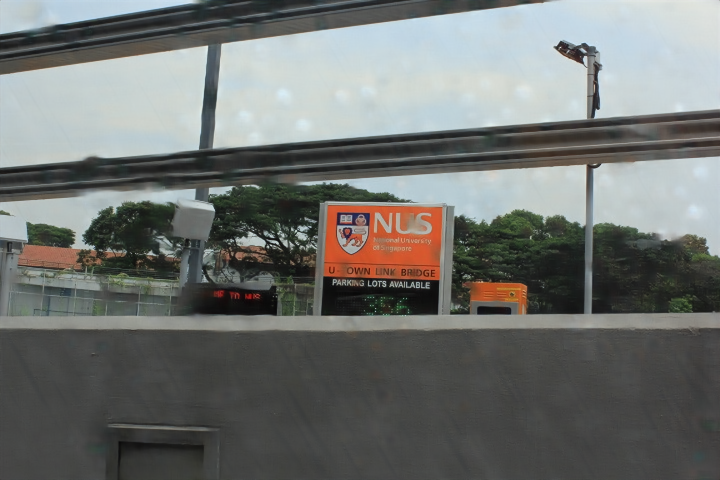}}
    \subfigure[Qian \textit{et al.} \cite{qian2018attentive}]{
    \label{jrdrs:c}
    \includegraphics[width=0.48\linewidth ]{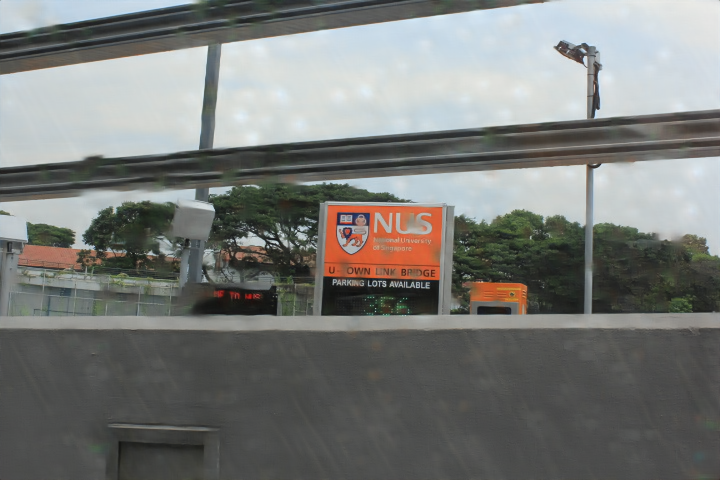}}
  \subfigure[Deraining: Ours]{
    \label{jrdrs:d}
    \includegraphics[width=0.48\linewidth ]{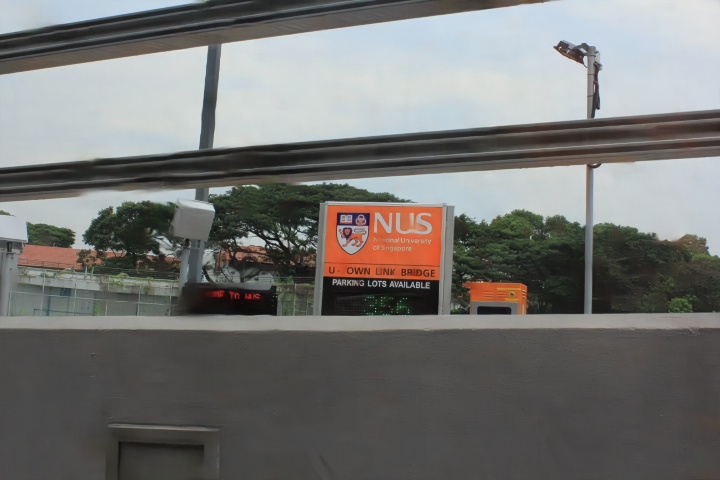}}
     \subfigure[Attention: rain streaks]{
    \label{jrdrs:e}
    \includegraphics[width=0.48\linewidth ]{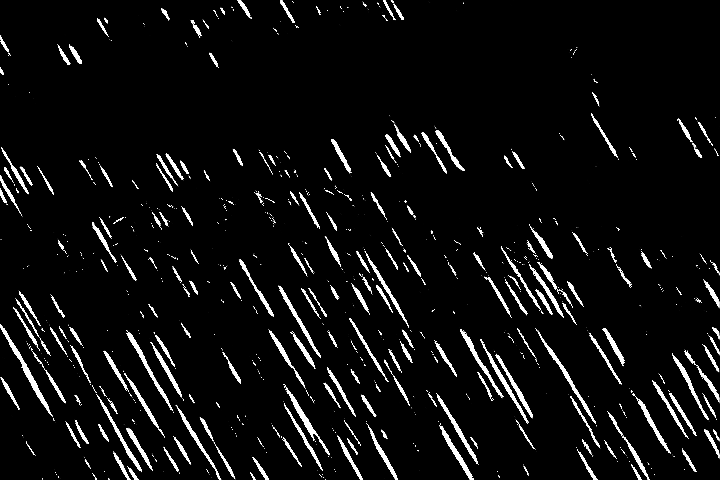}}
  \subfigure[Attention: raindrops]{
    \label{jrdrs:f}
    \includegraphics[width=0.48\linewidth ]{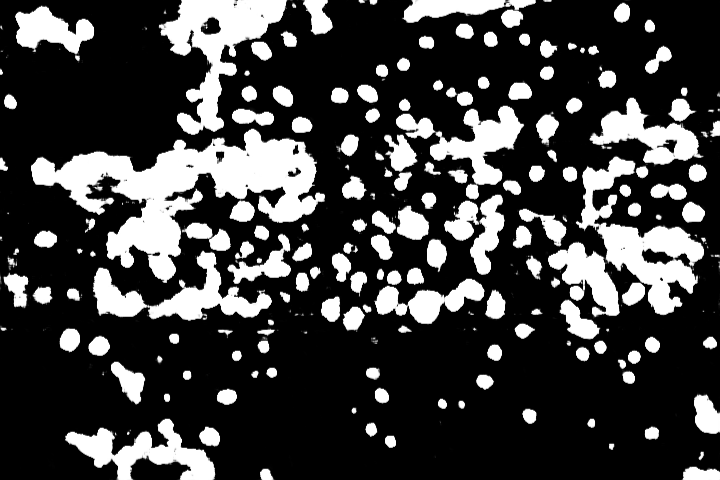}}
\caption{Rain streak and raindrop removal results on sample images from JRSRD dataset.}
  \label{figure_jrdrs}
\end{figure}

\begin{figure}[tb]
  \centering
  \subfigure[Input rainy image]{
    \label{ablation:a}
    \includegraphics[width=0.48\linewidth ]{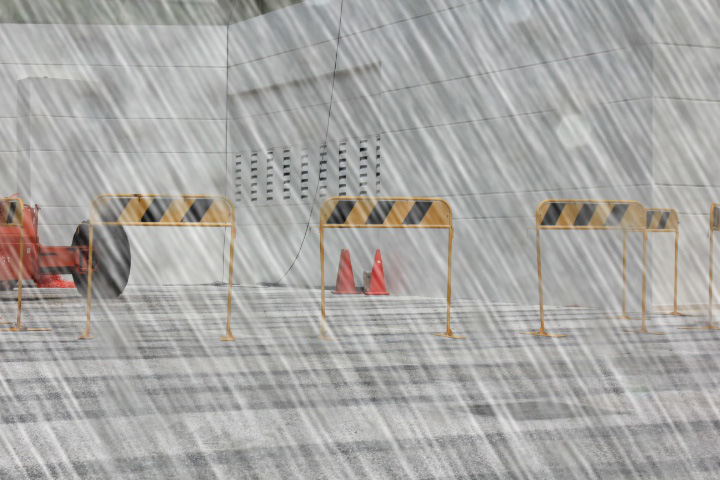}}
  \subfigure[DAM(odd)]{
    \label{ablation:b}
    \includegraphics[width=0.48\linewidth ]{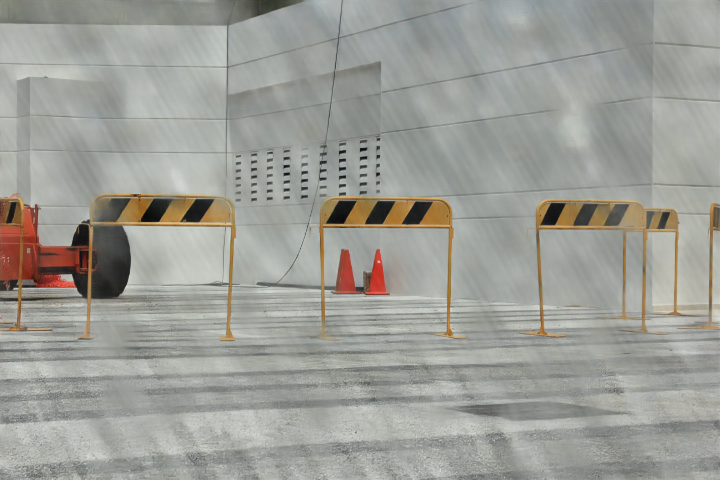}}
    \subfigure[DAM(dual)]{
    \label{ablation:c}
    \includegraphics[width=0.48\linewidth ]{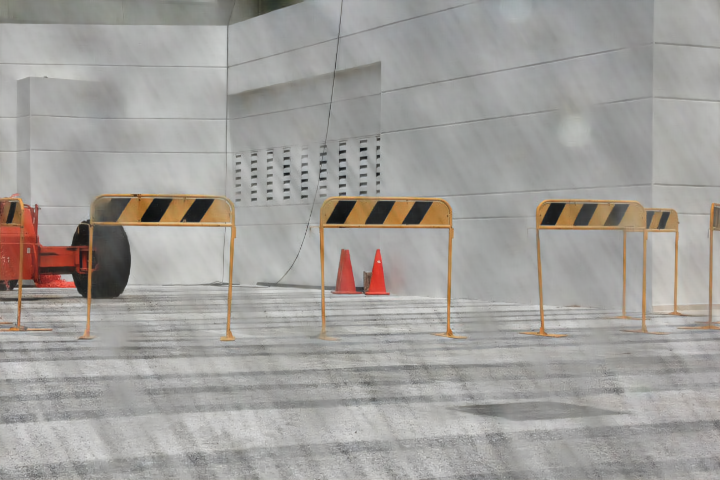}}
  \subfigure[DAiAM]{
    \label{ablation:d}
    \includegraphics[width=0.48\linewidth ]{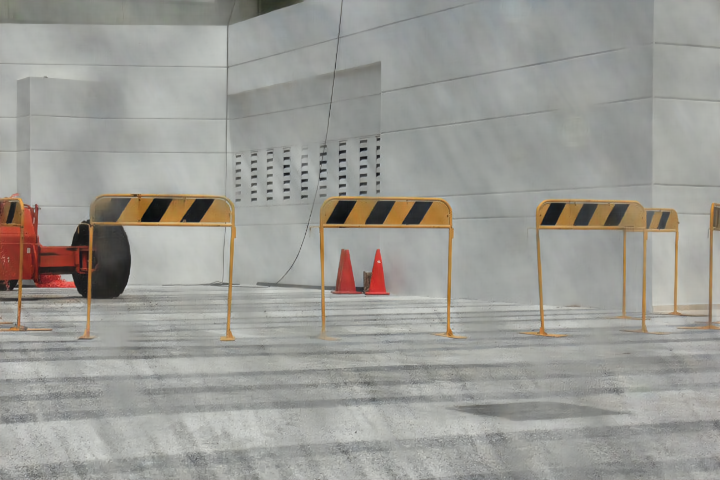}}
     \subfigure[DAiAM-DAiAM]{
    \label{ablation:e}
    \includegraphics[width=0.48\linewidth ]{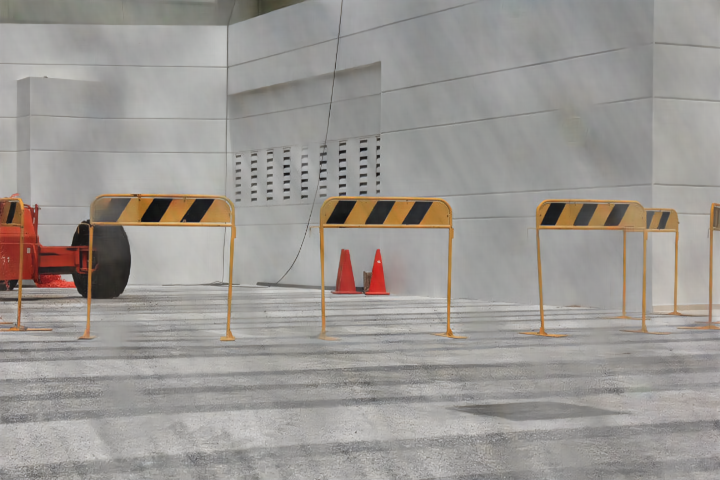}}
  \subfigure[D-DAiAM]{
    \label{ablation:f}
    \includegraphics[width=0.48\linewidth ]{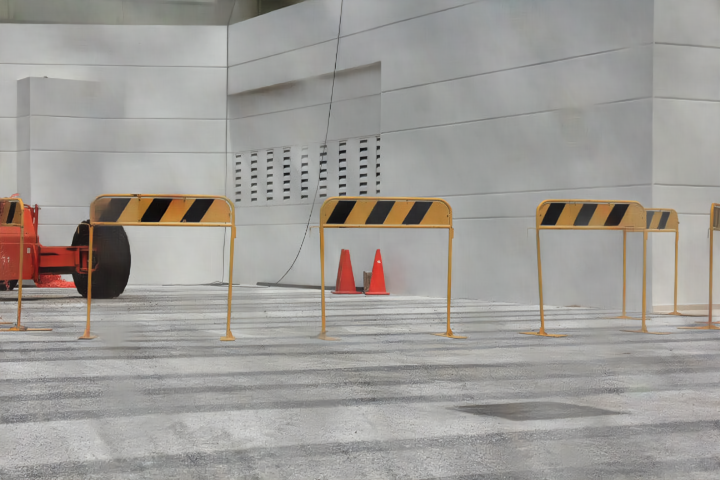}}
\caption{Ablation study results of rain streak and raindrop removal on sample images from JRSRD dataset. Zoom-in for details.}
  \label{figure_ablation}
\end{figure}

\begin{figure*}[tb]
  \centering
\includegraphics[width=0.97\linewidth ]{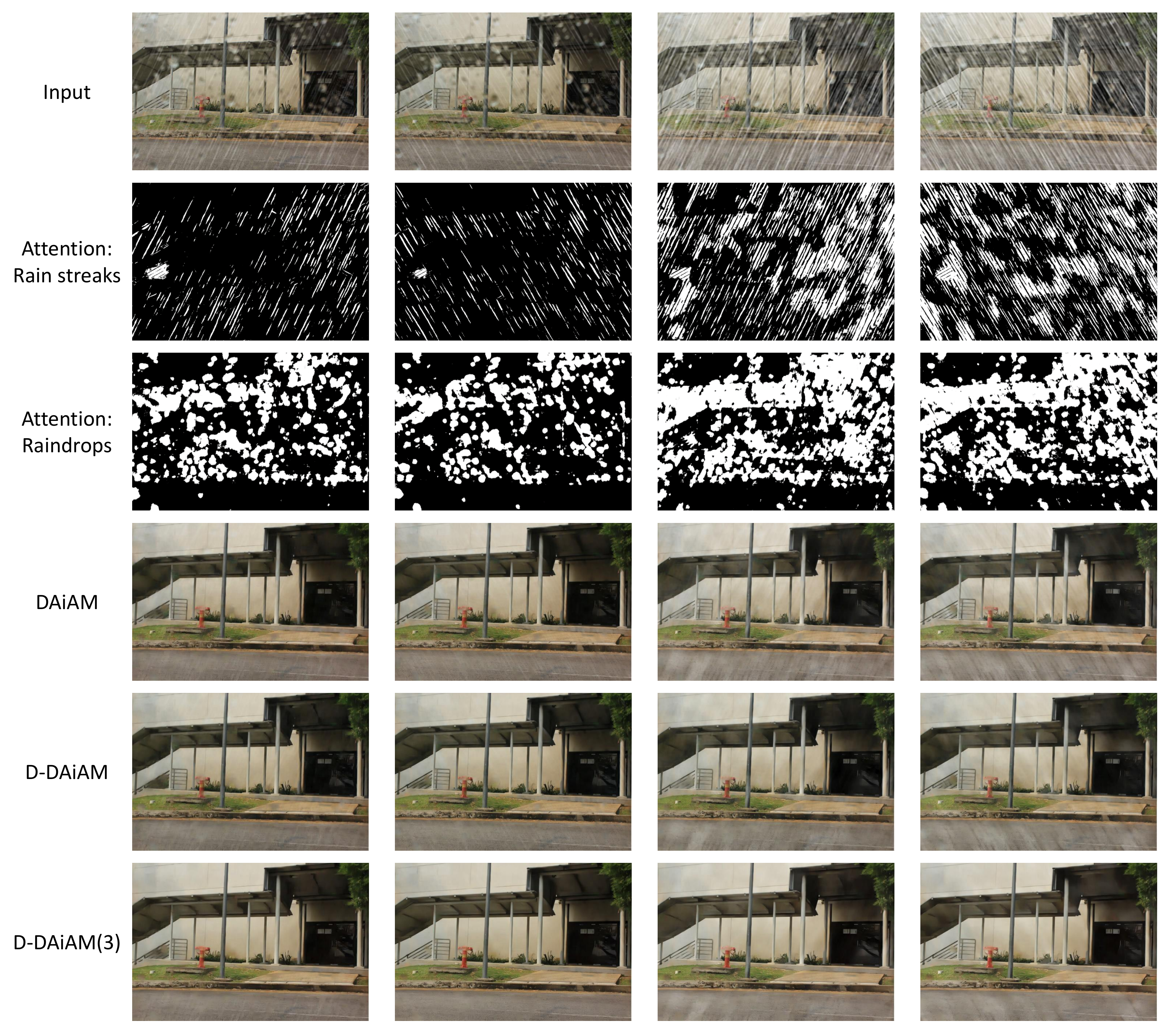}
\caption{Ablation study results of rain streaks and raindrop removal on different rain intensities. From top to bottom are the input, our attention maps for rain streaks and raindrops, DAiAM, D-DAiAM and D-DAiAM(3), respectively. Please zoom-in for details.}
\label{figure_intensity1}
\end{figure*}

\begin{figure*}[tb]
  \centering
\includegraphics[width=0.65\linewidth ]{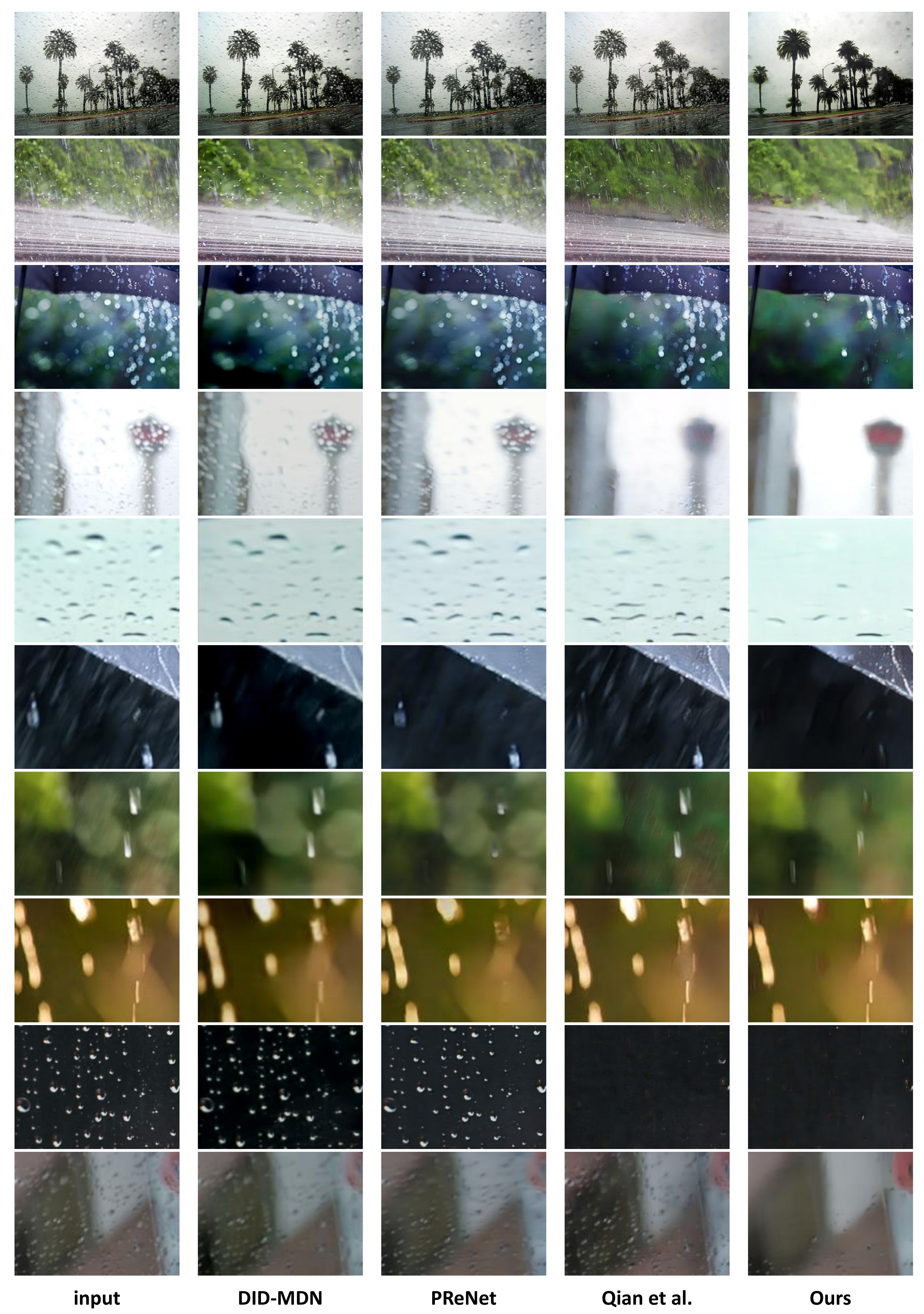}
\caption{The performance of different methods on real-world rainy images. From the left to right are the input, DID-MDN \cite{zhang2018density}, PReNet \cite{ren2019progressive}, Qian \textit{et al.} \cite{qian2018attentive} and ours. DID-MDN and PReNet are two rain streak removal methods, which only work on removing rain streaks. Qian \textit{et al.} is a raindrop removal method, which does not work on rain streak removal. Our proposed method achieves better performance by removing rain streaks and raindrops simultaneously on real-world rainy images.}
\label{figure_real}
\end{figure*}

\subsection{Results on the Joint Rain Streak and Raindrop Dataset}

There are many rain removal datasets for image deraining \cite{yang2017deep, qian2018attentive,zhang2018density,wang2019spatial,li2019single}. However, most of them focus on either rain streaks or raindrops. To this end, we synthesize a new joint rain streak and raindrop (JRSRD) dataset to evaluate the performance of different methods for removing both of them. Specially, the JRSRD training set contains $3,444$ synthetic rainy images, generated using images with raindrops from \cite{qian2018attentive}. We synthesize four images with different intensity levels of rain streaks for each image via Photoshop, which provides official methods to synthesize rain streak. In addition, many previous popular datasets are synthesized based on this strategy like Rain800 \cite{zhang2019image}, Rain12000 \cite{zhang2018density} and Rain14000 \cite{fu2017removing}. The noise levels are set between $20$\% and $60$\% to model various intensity. The JRSRD testing set contains $232$ pairs. The rainy images in our synthesized dataset contain both rain streaks and raindrops. Therefore, we apply DAiAM to remove rain. The performance compared with three current deraining methods is shown in Table \ref{table_jrdrs}. The ``Qian et al. \cite{qian2018attentive} + PReNet \cite{ren2019progressive}" means that we first use Qian et al. \cite{qian2018attentive} to remove raindrops from rainy images, and then use PReNet \cite{ren2019progressive} to remove rain streaks. The "PReNet \cite{ren2019progressive} + Qian et al. \cite{qian2018attentive}" represents the reverse order. Params means the parameters of different deep deraining networks. Time is the inference time. FLOPs means floating point operations.
All of them are re-trained on the proposed JRSRD dataset. Our proposed method beats these CNN-based methods on the task of joint rain streak and raindrop removal. Exemplar visual results are given in Fig. \ref{figure_jrdrs}, suggesting that the proposed method is capable of generating cleaner images.

\subsection{Ablation Study}

To demonstrate the effectiveness of DAM, DAiAM and differential-driven module, we compare these structures with several variant structures. Different from previous methods which merely focus on heavy rain, the proposed DAM generates two feature maps paying attention to heavy rain and light rain, respectively. Thus we compare to model without attention, DAM(zero), and the models with one or two attention maps, which are named as DAM(odd) and DAM(dual), respectively. Then, we compare the performance of the proposed dual attention-in-attention model, DAiAM, which can jointly perceive rain streaks and raindrops. The D-DAiAM is the model which removes rain using the differential-driven module. We compare it with the method directly connecting two DAiAM, termed as DAiAM-DAiAM. We also aggressively use two differential-driven modules in D-DAiAM(3). Table \ref{table_ablation} shows the performance of them in terms of PSNR and SSIM. Apparently, the counterpart without attention performs worst. Using attention of heavy rain improves the performance, as demonstrated by DAM(odd). While dual attention mechanism further improves the results. The DAiAM outperforms these three by simultaneously removing both raindrops and rain streaks. Directly connecting two DAiAM as DAiAM-DAiAM indeed boosts the values, while the improvement is not as significant as that of the proposed D-DAiAM. 
Fig. \ref{figure_ablation} presents exemplar visual deraining results, which also suggest the effectiveness of the proposed method.

\begin{table}[tb]
  \centering 
    \caption{Performance of different model structures on the JRSRD dataset.}
    \setlength\tabcolsep{5.0pt}
    \begin{tabular}{l |  c c c }
    \toprule
    Methods &  PSNR & SSIM & Params/Time/FLOPs\\
    \hline
    RESCAN \cite{li2018recurrent} & 21.05 & 0.768 & 0.15M/0.07s/2.46E+11 \\
    PReNet \cite{ren2019progressive} & 23.29 & 0.789 & 0.17M/0.10s/3.37E+11 \\
    Qian \textit{et al.} \cite{qian2018attentive} & 22.49 & 0.772 & 6.00M/0.13s/6.83E+11  \\
    \hline
    Qian \textit{et al.} \cite{qian2018attentive} + PReNet \cite{ren2019progressive}  & 23.89 & 0.796 &  6.17M/0.23s/10.2E+11  \\
    PReNet \cite{ren2019progressive} + Qian \textit{et al.} \cite{qian2018attentive} & 23.68 & 0.793 & 6.17M/0.23s/10.2E+11  \\
    \hline
    \textbf{DAiAM} & \textbf{24.67} & \textbf{0.819} &  3.60M/0.13s/8.90E+11 \\
    \textbf{D-DAiAM} & \textbf{25.26} & \textbf{0.825} & 7.20M/0.28s/17.8E+11  \\
    \bottomrule
    \end{tabular}%
    \label{table_jrdrs}
\end{table}%

\begin{table}[tb]
  \centering 
    \caption{Ablation study on the JRSRD dataset in terms of PSNR and SSIM.}
    \setlength\tabcolsep{5.0pt}
    \begin{tabular}{l |  c c }
    \toprule
    Methods &  PSNR & SSIM \\
    \hline
    DAM(zero) & 21.97 & 0.729 \\
    DAM(odd) & 23.41 & 0.791 \\
    DAM(dual) & 24.15 & 0.806  \\
    DAiAM & 24.67 & 0.819 \\
    \hline
    DAiAM-DAiAM & 24.84 & 0.823 \\
    D-DAiAM & 25.26 & 0.825  \\
    D-DAiAM(3) & 25.68 & 0.833  \\
    \bottomrule
    \end{tabular}%
    \label{table_ablation}
\end{table}%

\subsection{Performance of Deraining for Different Rain Intensities}

In this section, we evaluate the proposed method for removing different types of rain with different levels of intensity. Fig. \ref{figure_intensity1} present exemplar visual deraining results of our full method D-DAiAM and its two variants, DAiAM and D-DAiAM(3). Observing from the comparison, we have the following the findings. 1) It is more difficult to remove heavy rain than the light rain. All the methods exhibit more artifacts when then rain intensity becomes heavier. 2) Compared with DAiAM, our D-DAiAM and D-DAiAM(3) refine the performance of rain removal, especially for heavy rain, which shows the effectiveness of the differential-driven module. 3) The proposed method is capable of focusing on rain streaks and raindrops simultaneously, which shows the effectiveness of the dual attention-in-attention model. 4) Meanwhile, during the generation of attention maps, the rain streaks and raindrops can affect each other, which will cause erroneous attention maps. This shows the necessity of focusing on heavy-aware and light-aware regions simultaneously.

\subsection{Deployment in Real World} 
The proposed method is also evaluated on real-world images from the Internet. Fig. \ref{figure_real} shows the visual deraining results of different methods. DID-MDN \cite{zhang2018density} and PReNet \cite{ren2019progressive} are two state-of-the-art methods for rain streak removal, and Qian \textit{et al.} \cite{qian2018attentive} is one of the best methods to remove raindrops \cite{li2019single}. 
The proposed method achieves better performance on removing both rain streaks and raindrops than other methods, due to the proposed dual attention-in-attention mechanism. Rain streaks and raindrops are focused simultaneously via this mechanism-based network. Inside the proposed network, there are two well-designed DAMs, which also focus on local regions with different rainy intensities. The intensity-aware attention maps enable better removal of rain in different regions. The compared methods can only remove either raindrops (\textit{e.g.}, \cite{qian2018attentive}) or rain streaks (\textit{e.g.}, \cite{zhang2018density} and \cite{ren2019progressive}).
Consider that Fig. \ref{figure_real} has shown that the proposed method achieves significantly better results than other models, thus we do not conduct quantitative evaluation such as recognition score or user study.

\section{Conclusion}

In this paper, we tackle the problem of joint removal of raindrops and rain streaks. A dual attention-in-attention model, DAiAM, is presented to focus on raindrops and rain streaks simultaneously. Inside DAiAM, we propose a dual attention model, DAM. The proposed DAM learns two intensity-aware maps to remove rain from heavy and light rainy regions. We further introduce a differential-driven module to optimize the deraining process. Experimental results have demonstrated that our method performs best against the state-of-the-art methods and is capable of deraining well in real-world scenarios. In the future, we will consider generating images with purely heavy or purely light rains for supervision to help remove rains.

\section*{Acknowledgment}
This work is supported by Fund project of Jimei University (No. zp2020042), Xiamen Key Laboratory of Marine Intelligent Terminal R\&D and Application (No. B18208).

\bibliographystyle{IEEEtran}
\bibliography{egbib}

\begin{thebibliography}{10}
\providecommand{\url}[1]{#1}
\csname url@samestyle\endcsname
\providecommand{\newblock}{\relax}
\providecommand{\bibinfo}[2]{#2}
\providecommand{\BIBentrySTDinterwordspacing}{\spaceskip=0pt\relax}
\providecommand{\BIBentryALTinterwordstretchfactor}{4}
\providecommand{\BIBentryALTinterwordspacing}{\spaceskip=\fontdimen2\font plus
\BIBentryALTinterwordstretchfactor\fontdimen3\font minus
  \fontdimen4\font\relax}
\providecommand{\BIBforeignlanguage}[2]{{%
\expandafter\ifx\csname l@#1\endcsname\relax
\typeout{** WARNING: IEEEtran.bst: No hyphenation pattern has been}%
\typeout{** loaded for the language `#1'. Using the pattern for}%
\typeout{** the default language instead.}%
\else
\language=\csname l@#1\endcsname
\fi
#2}}
\providecommand{\BIBdecl}{\relax}
\BIBdecl

\bibitem{girshick2015fast}
R.~Girshick, ``Fast r-cnn,'' in \emph{Proceedings of the IEEE international
  conference on computer vision}, 2015.

\bibitem{he2017mask}
K.~He, G.~Gkioxari, P.~Doll{\'a}r, and R.~Girshick, ``Mask r-cnn,'' in
  \emph{Proceedings of the IEEE international conference on computer vision},
  2017.

\bibitem{zheng2015scalable}
L.~Zheng, L.~Shen, L.~Tian, S.~Wang, J.~Wang, and Q.~Tian, ``Scalable person
  re-identification: A benchmark,'' in \emph{Proceedings of the IEEE
  international conference on computer vision}, 2015.

\bibitem{han2005individual}
J.~Han and B.~Bhanu, ``Individual recognition using gait energy image,''
  \emph{IEEE transactions on pattern analysis and machine intelligence},
  vol.~28, no.~2, pp. 316--322, 2005.

\bibitem{yang2019drivingstereo}
G.~Yang, X.~Song, C.~Huang, Z.~Deng, J.~Shi, and B.~Zhou, ``Drivingstereo: A
  large-scale dataset for stereo matching in autonomous driving scenarios,'' in
  \emph{Proceedings of the IEEE Conference on Computer Vision and Pattern
  Recognition}, 2019.

\bibitem{li2019gs3d}
B.~Li, W.~Ouyang, L.~Sheng, X.~Zeng, and X.~Wang, ``Gs3d: An efficient 3d
  object detection framework for autonomous driving,'' in \emph{Proceedings of
  the IEEE Conference on Computer Vision and Pattern Recognition}, 2019.

\bibitem{sun2014exploiting}
S.-H. Sun, S.-P. Fan, and Y.-C.~F. Wang, ``Exploiting image structural
  similarity for single image rain removal,'' in \emph{IEEE International
  Conference on Image Processing}, 2014.

\bibitem{kang2011automatic}
L.-W. Kang, C.-W. Lin, and Y.-H. Fu, ``Automatic single-image-based rain
  streaks removal via image decomposition,'' \emph{IEEE transactions on image
  processing}, vol.~21, no.~4, pp. 1742--1755, 2011.

\bibitem{chen2013generalized}
Y.-L. Chen and C.-T. Hsu, ``A generalized low-rank appearance model for
  spatio-temporally correlated rain streaks,'' in \emph{Proceedings of the IEEE
  international conference on computer vision}, 2013.

\bibitem{zhang2006rain}
X.~Zhang, H.~Li, Y.~Qi, W.~K. Leow, and T.~K. Ng, ``Rain removal in video by
  combining temporal and chromatic properties,'' in \emph{2006 IEEE
  international conference on multimedia and expo}.\hskip 1em plus 0.5em minus
  0.4em\relax IEEE, 2006, pp. 461--464.

\bibitem{fu2017clearing}
X.~Fu, J.~Huang, X.~Ding, Y.~Liao, and J.~Paisley, ``Clearing the skies: A deep
  network architecture for single-image rain removal,'' \emph{IEEE Transactions
  on Image Processing}, vol.~26, no.~6, pp. 2944--2956, 2017.

\bibitem{fu2017removing}
X.~Fu, J.~Huang, D.~Zeng, Y.~Huang, X.~Ding, and J.~Paisley, ``Removing rain
  from single images via a deep detail network,'' in \emph{Proceedings of the
  IEEE Conference on Computer Vision and Pattern Recognition}, 2017.

\bibitem{li2018recurrent}
X.~Li, J.~Wu, Z.~Lin, H.~Liu, and H.~Zha, ``Recurrent squeeze-and-excitation
  context aggregation net for single image deraining,'' in \emph{European
  Conference on Computer Vision}, 2018.

\bibitem{yang2017deep}
W.~Yang, R.~T. Tan, J.~Feng, J.~Liu, Z.~Guo, and S.~Yan, ``Deep joint rain
  detection and removal from a single image,'' in \emph{Proceedings of the IEEE
  Conference on Computer Vision and Pattern Recognition}, 2017.

\bibitem{zhang2018density}
H.~Zhang and V.~M. Patel, ``Density-aware single image de-raining using a
  multi-stream dense network,'' in \emph{Proceedings of the IEEE Conference on
  Computer Vision and Pattern Recognition}, 2018.

\bibitem{roser2009video}
M.~Roser and A.~Geiger, ``Video-based raindrop detection for improved image
  registration,'' in \emph{Proceedings of the IEEE international conference on
  computer vision Workshops}, 2009.

\bibitem{roser2010realistic}
M.~Roser, J.~Kurz, and A.~Geiger, ``Realistic modeling of water droplets for
  monocular adherent raindrop recognition using bezier curves,'' in \emph{Asian
  Conference on Computer Vision}, 2010.

\bibitem{eigen2013restoring}
D.~Eigen, D.~Krishnan, and R.~Fergus, ``Restoring an image taken through a
  window covered with dirt or rain,'' in \emph{Proceedings of the IEEE
  international conference on computer vision}, 2013.

\bibitem{qian2018attentive}
R.~Qian, R.~T. Tan, W.~Yang, J.~Su, and J.~Liu, ``Attentive generative
  adversarial network for raindrop removal from a single image,'' in
  \emph{Proceedings of the IEEE Conference on Computer Vision and Pattern
  Recognition}, 2018.

\bibitem{barnum2010analysis}
P.~C. Barnum, S.~Narasimhan, and T.~Kanade, ``Analysis of rain and snow in
  frequency space,'' \emph{International journal of computer vision}, vol.~86,
  no.~2, pp. 256--274, 2010.

\bibitem{huang2013self}
D.-A. Huang, L.-W. Kang, Y.-C.~F. Wang, and C.-W. Lin, ``Self-learning based
  image decomposition with applications to single image denoising,'' \emph{IEEE
  Transactions on multimedia}, vol.~16, no.~1, pp. 83--93, 2013.

\bibitem{luo2015removing}
Y.~Luo, Y.~Xu, and H.~Ji, ``Removing rain from a single image via
  discriminative sparse coding,'' in \emph{Proceedings of the IEEE
  International Conference on Computer Vision}, 2015, pp. 3397--3405.

\bibitem{li2016rain}
Y.~Li, R.~T. Tan, X.~Guo, J.~Lu, and M.~S. Brown, ``Rain streak removal using
  layer priors,'' in \emph{Proceedings of the IEEE Conference on Computer
  Vision and Pattern Recognition}, 2016.

\bibitem{chang2017transformed}
Y.~Chang, L.~Yan, and S.~Zhong, ``Transformed low-rank model for line pattern
  noise removal,'' in \emph{Proceedings of the IEEE international conference on
  computer vision}, 2017.

\bibitem{zhu2017joint}
L.~Zhu, C.-W. Fu, D.~Lischinski, and P.-A. Heng, ``Joint bi-layer optimization
  for single-image rain streak removal,'' in \emph{Proceedings of the IEEE
  international conference on computer vision}, 2017.

\bibitem{zhu2020learning}
L.~Zhu, Z.~Deng, X.~Hu, H.~Xie, X.~Xu, J.~Qin, and P.-A. Heng, ``Learning gated
  non-local residual for single-image rain streak removal,'' \emph{IEEE
  Transactions on Circuits and Systems for Video Technology}, 2020.

\bibitem{hu2021single}
X.~Hu, L.~Zhu, T.~Wang, C.-W. Fu, and P.-A. Heng, ``Single-image real-time rain
  removal based on depth-guided non-local features,'' \emph{IEEE Transactions
  on Image Processing}, vol.~30, pp. 1759--1770, 2021.

\bibitem{wang2020rethinking}
Y.~Wang, Y.~Song, C.~Ma, and B.~Zeng, ``Rethinking image deraining via rain
  streaks and vapors,'' in \emph{European Conference on Computer Vision}.\hskip
  1em plus 0.5em minus 0.4em\relax Springer, 2020, pp. 367--382.

\bibitem{johnson2016perceptual}
J.~Johnson, A.~Alahi, and L.~Fei-Fei, ``Perceptual losses for real-time style
  transfer and super-resolution,'' in \emph{European conference on computer
  vision}.\hskip 1em plus 0.5em minus 0.4em\relax Springer, 2016, pp. 694--711.

\bibitem{zhang2018adversarial}
K.~Zhang, W.~Luo, Y.~Zhong, L.~Ma, W.~Liu, and H.~Li, ``Adversarial
  spatio-temporal learning for video deblurring,'' \emph{IEEE Transactions on
  Image Processing}, vol.~28, no.~1, pp. 291--301, 2018.

\bibitem{zhang2020deblurring}
K.~Zhang, W.~Luo, Y.~Zhong, L.~Ma, B.~Stenger, W.~Liu, and H.~Li, ``Deblurring
  by realistic blurring,'' in \emph{Proceedings of the IEEE/CVF Conference on
  Computer Vision and Pattern Recognition}, 2020, pp. 2737--2746.

\bibitem{zheng2021t}
L.~Zheng, Y.~Li, K.~Zhang, and W.~Luo, ``T-net: Deep stacked scale-iteration
  network for image dehazing,'' \emph{arXiv preprint arXiv:2106.02809}, 2021.

\bibitem{zhang2020every}
K.~Zhang, W.~Luo, B.~Stenger, W.~Ren, L.~Ma, and H.~Li, ``Every moment matters:
  Detail-aware networks to bring a blurry image alive,'' in \emph{Proceedings
  of the 28th ACM International Conference on Multimedia}, 2020, pp. 384--392.

\bibitem{zhang2021deep}
K.~Zhang, R.~Li, Y.~Yu, W.~Luo, C.~Li, and H.~Li, ``Deep dense multi-scale
  network for snow removal using semantic and geometric priors,'' \emph{arXiv
  preprint arXiv:2103.11298}, 2021.

\bibitem{zhang2019image}
H.~Zhang, V.~Sindagi, and V.~M. Patel, ``Image de-raining using a conditional
  generative adversarial network,'' \emph{IEEE transactions on circuits and
  systems for video technology}, vol.~30, no.~11, pp. 3943--3956, 2019.

\bibitem{zhang2021beyond}
K.~Zhang, W.~Luo, Y.~Yu, W.~Ren, F.~Zhao, C.~Li, L.~Ma, W.~Liu, and H.~Li,
  ``Beyond monocular deraining: Parallel stereo deraining network via semantic
  prior,'' \emph{arXiv preprint arXiv:2105.03830}, 2021.

\bibitem{zhang2020beyond}
K.~Zhang, W.~Luo, W.~Ren, J.~Wang, F.~Zhao, L.~Ma, and H.~Li, ``Beyond
  monocular deraining: Stereo image deraining via semantic understanding,'' in
  \emph{European Conference on Computer Vision}.\hskip 1em plus 0.5em minus
  0.4em\relax Springer, 2020, pp. 71--89.

\bibitem{wang2020model}
H.~Wang, Q.~Xie, Q.~Zhao, and D.~Meng, ``A model-driven deep neural network for
  single image rain removal,'' in \emph{Proceedings of the IEEE Conference on
  Computer Vision and Pattern Recognition}, 2020.

\bibitem{li2021comprehensive}
S.~Li, W.~Ren, F.~Wang, I.~B. Araujo, E.~K. Tokuda, R.~H. Junior, R.~M.
  Cesar-Jr, Z.~Wang, and X.~Cao, ``A comprehensive benchmark analysis of single
  image deraining: Current challenges and future perspectives,''
  \emph{International Journal of Computer Vision}, vol. 129, no.~4, pp.
  1301--1322, 2021.

\bibitem{li2018video}
M.~Li, Q.~Xie, Q.~Zhao, W.~Wei, S.~Gu, J.~Tao, and D.~Meng, ``Video rain streak
  removal by multiscale convolutional sparse coding,'' in \emph{Proceedings of
  the IEEE Conference on Computer Vision and Pattern Recognition}, 2018.

\bibitem{liu2018d3r}
J.~Liu, W.~Yang, S.~Yang, and Z.~Guo, ``D3r-net: Dynamic routing residue
  recurrent network for video rain removal,'' \emph{IEEE Transactions on Image
  Processing}, vol.~28, no.~2, pp. 699--712, 2018.

\bibitem{liu2018erase}
------, ``Erase or fill? deep joint recurrent rain removal and reconstruction
  in videos,'' in \emph{Proceedings of the IEEE Conference on Computer Vision
  and Pattern Recognition}, 2018.

\bibitem{chen2018robust}
J.~Chen, C.-H. Tan, J.~Hou, L.-P. Chau, and H.~Li, ``Robust video content
  alignment and compensation for rain removal in a cnn framework,'' in
  \emph{Proceedings of the IEEE Conference on Computer Vision and Pattern
  Recognition}, 2018.

\bibitem{yang2019frame}
W.~Yang, J.~Liu, and J.~Feng, ``Frame-consistent recurrent video deraining with
  dual-level flow,'' in \emph{Proceedings of the IEEE Conference on Computer
  Vision and Pattern Recognition}, 2019.

\bibitem{zhang2021enhanced}
K.~Zhang, D.~Li, W.~Luo, W.-Y. Lin, F.~Zhao, W.~Ren, W.~Liu, and H.~Li,
  ``Enhanced spatio-temporal interaction learning for video deraining: A faster
  and better framework,'' \emph{arXiv preprint arXiv:2103.12318}, 2021.

\bibitem{kurihata2005rainy}
H.~Kurihata, T.~Takahashi, I.~Ide, Y.~Mekada, H.~Murase, Y.~Tamatsu, and
  T.~Miyahara, ``Rainy weather recognition from in-vehicle camera images for
  driver assistance,'' in \emph{IEEE Proceedings. Intelligent Vehicles
  Symposium, 2005.}\hskip 1em plus 0.5em minus 0.4em\relax IEEE, 2005, pp.
  205--210.

\bibitem{yamashita2005removal}
A.~Yamashita, Y.~Tanaka, and T.~Kaneko, ``Removal of adherent waterdrops from
  images acquired with stereo camera,'' in \emph{2005 IEEE/RSJ International
  Conference on Intelligent Robots and Systems}.\hskip 1em plus 0.5em minus
  0.4em\relax IEEE, 2005, pp. 400--405.

\bibitem{yamashita2009noises}
A.~Yamashita, I.~Fukuchi, and T.~Kaneko, ``Noises removal from image sequences
  acquired with moving camera by estimating camera motion from spatio-temporal
  information,'' in \emph{2009 IEEE/RSJ International Conference on Intelligent
  Robots and Systems}.\hskip 1em plus 0.5em minus 0.4em\relax IEEE, 2009, pp.
  3794--3801.

\bibitem{you2015adherent}
S.~You, R.~T. Tan, R.~Kawakami, Y.~Mukaigawa, and K.~Ikeuchi, ``Adherent
  raindrop modeling, detectionand removal in video,'' \emph{IEEE transactions
  on pattern analysis and machine intelligence}, vol.~38, no.~9, pp.
  1721--1733, 2015.

\bibitem{quan2019deep}
Y.~Quan, S.~Deng, Y.~Chen, and H.~Ji, ``Deep learning for seeing through window
  with raindrops,'' in \emph{Proceedings of the IEEE International Conference
  on Computer Vision}, 2019.

\bibitem{alletto2019adherent}
S.~Alletto, C.~Carlin, L.~Rigazio, Y.~Ishii, and S.~Tsukizawa, ``Adherent
  raindrop removal with self-supervised attention maps and spatio-temporal
  generative adversarial networks,'' in \emph{Proceedings of the IEEE/CVF
  International Conference on Computer Vision Workshops}, 2019, pp. 0--0.

\bibitem{hao2019learning}
Z.~Hao, S.~You, Y.~Li, K.~Li, and F.~Lu, ``Learning from synthetic
  photorealistic raindrop for single image raindrop removal,'' in
  \emph{Proceedings of the IEEE/CVF International Conference on Computer Vision
  Workshops}, 2019, pp. 0--0.

\bibitem{li2019single}
S.~Li, I.~B. Araujo, W.~Ren, Z.~Wang, E.~K. Tokuda, R.~H. Junior,
  R.~Cesar-Junior, J.~Zhang, X.~Guo, and X.~Cao, ``Single image deraining: A
  comprehensive benchmark analysis,'' in \emph{Proceedings of the IEEE
  Conference on Computer Vision and Pattern Recognition}, 2019.

\bibitem{ba2014multiple}
J.~Ba, V.~Mnih, and K.~Kavukcuoglu, ``Multiple object recognition with visual
  attention,'' \emph{arXiv preprint arXiv:1412.7755}, 2014.

\bibitem{gregor2015draw}
K.~Gregor, I.~Danihelka, A.~Graves, D.~J. Rezende, and D.~Wierstra, ``Draw: A
  recurrent neural network for image generation,'' \emph{arXiv preprint
  arXiv:1502.04623}, 2015.

\bibitem{xiao2015application}
T.~Xiao, Y.~Xu, K.~Yang, J.~Zhang, Y.~Peng, and Z.~Zhang, ``The application of
  two-level attention models in deep convolutional neural network for
  fine-grained image classification,'' in \emph{Proceedings of the IEEE
  Conference on Computer Vision and Pattern Recognition}, 2015.

\bibitem{xu2015show}
K.~Xu, J.~Ba, R.~Kiros, K.~Cho, A.~Courville, R.~Salakhudinov, R.~Zemel, and
  Y.~Bengio, ``Show, attend and tell: Neural image caption generation with
  visual attention,'' in \emph{International Conference on Machine Learning},
  2015.

\bibitem{you2016image}
Q.~You, H.~Jin, Z.~Wang, C.~Fang, and J.~Luo, ``Image captioning with semantic
  attention,'' in \emph{Proceedings of the IEEE Conference on Computer Vision
  and Pattern Recognition}, 2016.

\bibitem{chen2018reverse}
S.~Chen, X.~Tan, B.~Wang, and X.~Hu, ``Reverse attention for salient object
  detection,'' in \emph{European Conference on Computer Vision}, 2018.

\bibitem{lv2021simultaneously}
Y.~Lv, J.~Zhang, Y.~Dai, A.~Li, B.~Liu, N.~Barnes, and D.-P. Fan,
  ``Simultaneously localize, segment and rank the camouflaged objects,'' in
  \emph{Proceedings of the IEEE/CVF Conference on Computer Vision and Pattern
  Recognition}, 2021, pp. 11\,591--11\,601.

\bibitem{li2021uncertainty}
A.~Li, J.~Zhang, Y.~Lv, B.~Liu, T.~Zhang, and Y.~Dai, ``Uncertainty-aware joint
  salient object and camouflaged object detection,'' in \emph{Proceedings of
  the IEEE/CVF Conference on Computer Vision and Pattern Recognition}, 2021,
  pp. 10\,071--10\,081.

\bibitem{mao2021transformer}
Y.~Mao, J.~Zhang, Z.~Wan, Y.~Dai, A.~Li, Y.~Lv, X.~Tian, D.-P. Fan, and
  N.~Barnes, ``Transformer transforms salient object detection and camouflaged
  object detection,'' \emph{arXiv preprint arXiv:2104.10127}, 2021.

\bibitem{wang2019spatial}
T.~Wang, X.~Yang, K.~Xu, S.~Chen, Q.~Zhang, and R.~W. Lau, ``Spatial attentive
  single-image deraining with a high quality real rain dataset,'' in
  \emph{Proceedings of the IEEE Conference on Computer Vision and Pattern
  Recognition}, 2019.

\bibitem{hu2019depth}
X.~Hu, C.-W. Fu, L.~Zhu, and P.-A. Heng, ``Depth-attentional features for
  single-image rain removal,'' in \emph{Proceedings of the IEEE Conference on
  Computer Vision and Pattern Recognition}, 2019.

\bibitem{li2019heavy}
R.~Li, L.-F. Cheong, and R.~T. Tan, ``Heavy rain image restoration: Integrating
  physics model and conditional adversarial learning,'' in \emph{Proceedings of
  the IEEE Conference on Computer Vision and Pattern Recognition}, 2019.

\bibitem{ren2019progressive}
D.~Ren, W.~Zuo, Q.~Hu, P.~Zhu, and D.~Meng, ``Progressive image deraining
  networks: a better and simpler baseline,'' in \emph{Proceedings of the IEEE
  Conference on Computer Vision and Pattern Recognition}, 2019.

\bibitem{he2016deep}
K.~He, X.~Zhang, S.~Ren, and J.~Sun, ``Deep residual learning for image
  recognition,'' in \emph{Proceedings of the IEEE Conference on Computer Vision
  and Pattern Recognition}, 2016.

\bibitem{garg2006photorealistic}
K.~Garg and S.~K. Nayar, ``Photorealistic rendering of rain streaks,''
  \emph{ACM Transactions on Graphics (TOG)}, vol.~25, no.~3, pp. 996--1002,
  2006.

\end{thebibliography}
%

\begin{IEEEbiography}[{\includegraphics[width=1in,height=1.25in,clip,keepaspectratio]{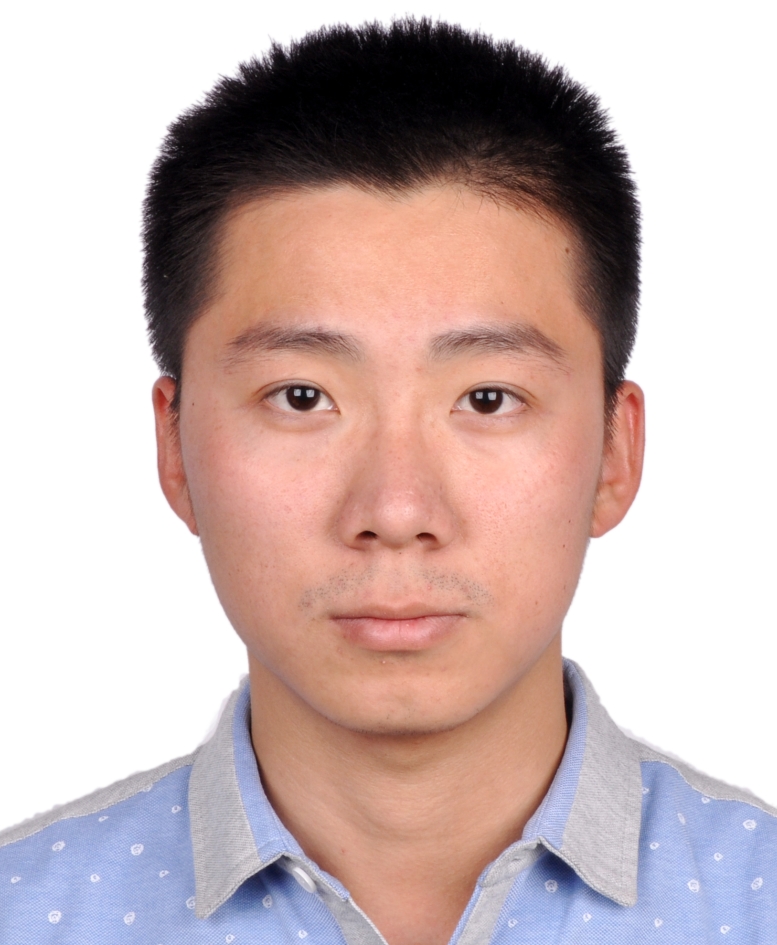}}]{Kaihao Zhang}
is currently pursuing the Ph.D. degree with the College of Engineering and Computer Science, The Australian National University, Canberra, ACT, Australia. His research interests focus on computer vision and deep learning. He has more than 20 referred publications in international conferences and journals, including CVPR, ICCV, ECCV, NeurIPS, AAAI, ACMMM, IJCV, TIP, TMM, etc. 
\end{IEEEbiography}

\begin{IEEEbiography}[{\includegraphics[width=1in,height=1.25in,clip,keepaspectratio]{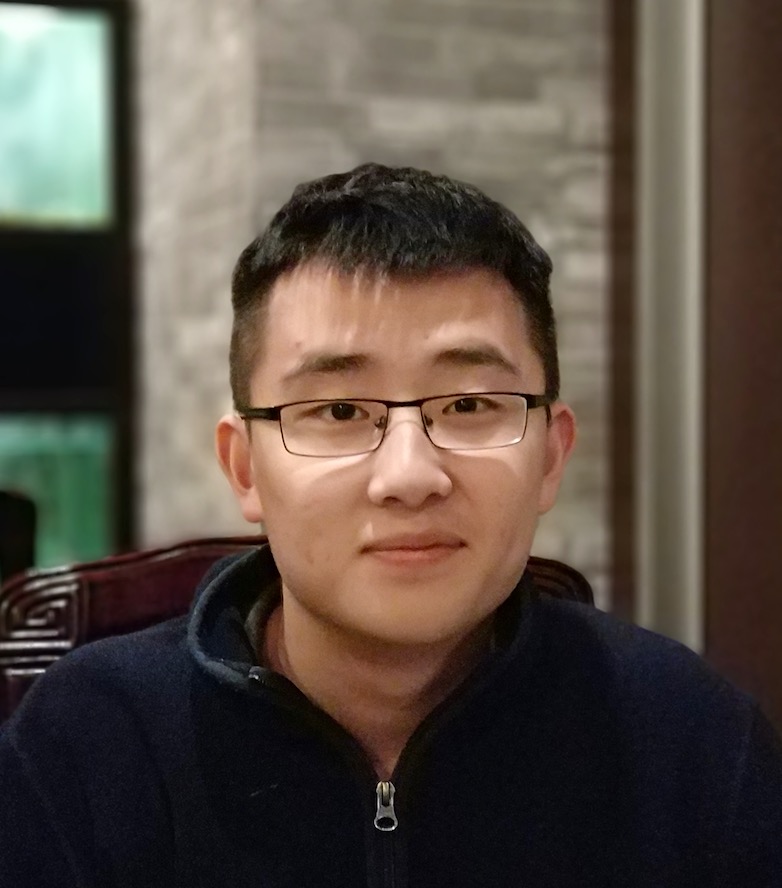}}]{Dongxu Li}
is a Ph.D candidate at The Australian National University. His research interests are mainly computer vision and deep learning, including visual sequence representation learning, vision-language learning and multi-modal learning. Before starting PhD, Dongxu obtained his Bachelor degree from The Australian National University with first-class honours in Computing.
\end{IEEEbiography}

\begin{IEEEbiography}[{\includegraphics[width=1in,height=1.25in,clip,keepaspectratio]{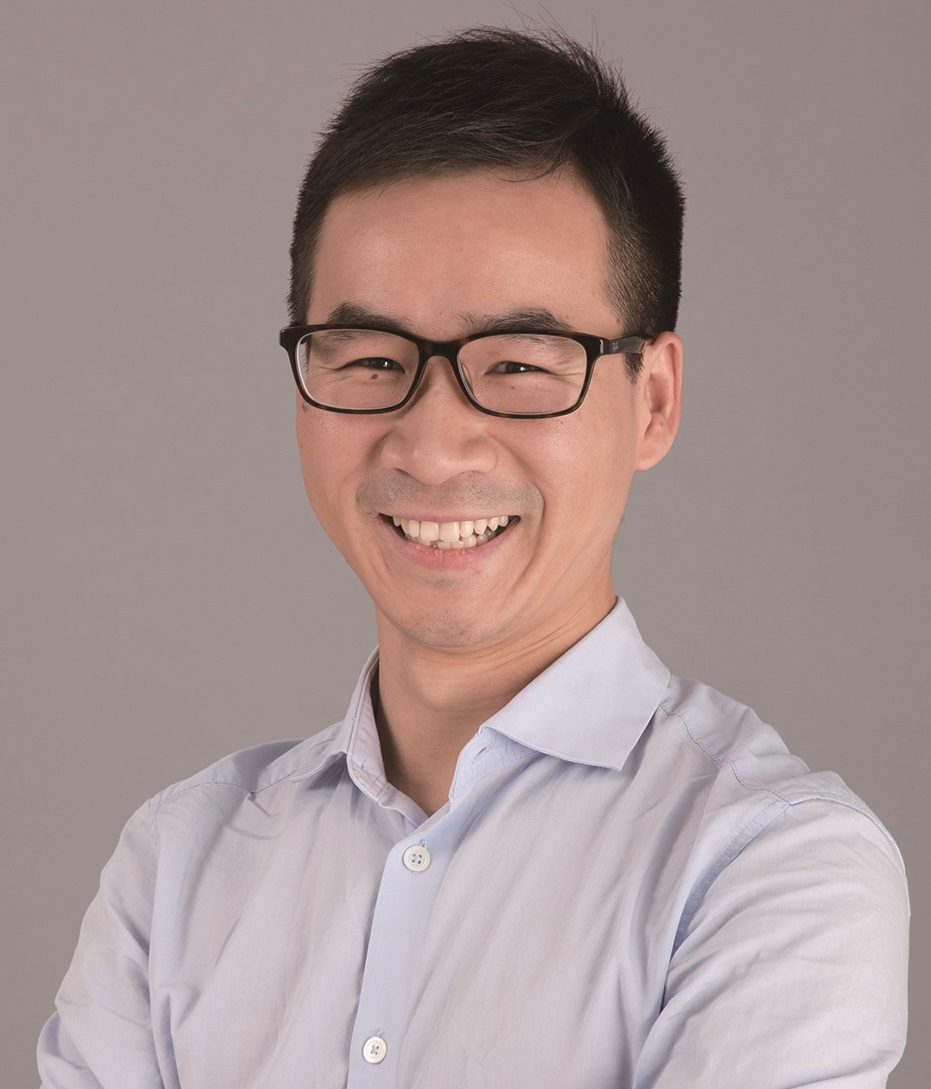}}]{Wenhan Luo}
is currently working as a senior researcher in the Tencent, China. His research interests include several topics in computer vision and machine learning, such as motion analysis (especially object tracking), image/video quality restoration, reinforcement learning. Before joining Tencent, he received the Ph.D. degree from Imperial College London, UK, 2016, M.E. degree from Institute of Automation, Chinese Academy of Sciences, China, 2012 and B.E. degree from Huazhong University of Science and Technology, China, 2009.
\end{IEEEbiography}

\begin{IEEEbiography}[{\includegraphics[width=1in,height=1.25in,clip,keepaspectratio]{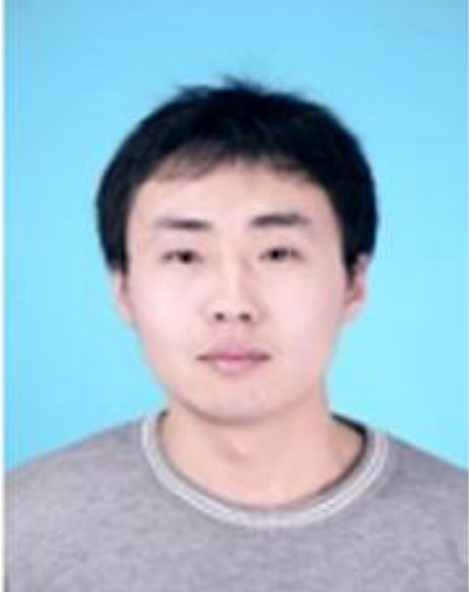}}]{Wenqi Ren} is an Associate Professor in Institute of Information Engineering, Chinese Academy of Sciences, China. He received his Ph.D. degree from Tianjin University, Tianjin, China, in 2017. During 2015 to 2016, he was supported by China Scholarship Council and working with Prof. Ming-Husan Yang as a joint-training Ph.D. student in the Electrical Engineering and Computer Science Department, at the University of California at Merced. He received Tencent Rhino Bird Elite Graduate Program Scholarship in 2017, MSRA Star Track Program in 2018. His research interests include image processing and related high-level vision problems. 
\end{IEEEbiography}




\end{document}